\pdfoutput=1
\documentclass{article}

\usepackage{arxiv}

\usepackage[utf8]{inputenc}
\usepackage[T1]{fontenc}
\usepackage{mathrsfs}
\usepackage{graphicx}
\usepackage{multirow}
\usepackage{amsmath,amssymb,amsfonts}
\usepackage{amsthm}
\usepackage[title]{appendix}
\usepackage{xcolor}
\usepackage{textcomp}
\usepackage{booktabs}
\usepackage{listings}
\usepackage{longtable}
\usepackage{array}
\usepackage{float}
\usepackage{pifont}
\usepackage{siunitx}
\usepackage{colortbl}
\usepackage[numbers,sort&compress]{natbib}
\usepackage{url}
\usepackage[hidelinks]{hyperref}
\usepackage{orcidlink}

\ifx\endbibitem\undefined\def\endbibitem{}\fi

\definecolor{openeocubes}{RGB}{232, 242, 255}
\definecolor{daskml}{RGB}{232, 255, 236}

\newcommand{\cmark}{\ding{51}}
\newcommand{\xmark}{\ding{55}}

\definecolor{notebg}{gray}{0.9}
\newlength{\Lnote}

\definecolor{lstbackground}{RGB}{252, 251, 248}
\definecolor{lstcomment}   {RGB}{ 40, 140,  70}
\definecolor{lstkeyword}   {RGB}{ 30, 120, 200}
\definecolor{lstkeywordB}  {RGB}{160,  50, 170}
\definecolor{lstkeywordC}  {RGB}{190, 110,  10}
\definecolor{lststring}    {RGB}{180,  40,  20}
\definecolor{lstfunction}  {RGB}{ 10, 120, 140}
\definecolor{lstnumber}    {RGB}{150,  60,  60}
\definecolor{lstlinenumber}{RGB}{160, 158, 150}
\definecolor{lstrulecolor} {RGB}{210, 208, 200}
\definecolor{lstdollar}    {RGB}{ 80,  80,  80}
\definecolor{lstobject}    {RGB}{190,  80,   0}

\lstdefinestyle{basestyle}{
    backgroundcolor  = \color{lstbackground},
    commentstyle     = \color{lstcomment}\itshape,
    stringstyle      = \color{lststring}\bfseries,
    numberstyle      = \tiny\color{lstlinenumber},
    basicstyle       = \ttfamily\footnotesize,
    breakatwhitespace= false,
    breaklines       = true,
    captionpos       = b,
    keepspaces       = true,
    numbers          = left,
    stepnumber       = 1,
    numbersep        = 8pt,
    showspaces       = false,
    showstringspaces = false,
    showtabs         = false,
    tabsize          = 2,
    columns          = flexible,
    frame            = single,
    framesep         = 4pt,
    rulecolor        = \color{lstrulecolor},
    xleftmargin      = 16pt,
    xrightmargin     = 4pt,
    framexleftmargin = 16pt,
    aboveskip        = \medskipamount,
    belowskip        = \medskipamount,
    mathescape       = false,
    literate         =
        {p\$}{{\textcolor{lstobject}{p}\textcolor{lstdollar}{\$}}}2
        {dc\$}{{\textcolor{lstobject}{dc}\textcolor{lstdollar}{\$}}}3
        {con\$}{{\textcolor{lstobject}{con}\textcolor{lstdollar}{\$}}}4
        {\$}{{\textcolor{lstdollar}{\$}}}1,
}

\lstdefinestyle{python}{
    style            = basestyle,
    language         = Python,
    keywordstyle     = \color{lstkeyword}\bfseries,
    morekeywords     = [2]{
        True, False, None,
        abs, all, any, ascii, bin, bool, breakpoint, bytearray,
        bytes, callable, chr, classmethod, compile, complex,
        copyright, credits, delattr, dict, dir, divmod,
        enumerate, eval, exec, exit, filter, float, format,
        frozenset, getattr, globals, hasattr, hash, help,
        hex, id, input, int, isinstance, issubclass, iter,
        len, license, list, locals, map, max, memoryview,
        min, next, object, oct, open, ord, pow, print,
        property, quit, range, repr, reversed, round,
        set, setattr, slice, sorted, staticmethod, str,
        sum, super, tuple, type, vars, zip,
        __init__, __str__, __repr__, __len__, __call__,
        __name__, __main__, __file__, __doc__
    },
    keywordstyle     = [2]\color{lstkeywordB}\bfseries,
    morekeywords     = [3]{
        import, from, as,
        numpy, pandas, geopandas, rasterio, xarray,
        torch, tensorflow, keras, sklearn, onnx,
        onnxruntime, gdal, osgeo, shapely, pyproj,
        pystac, pydantic, requests, json, os, sys,
        pathlib, datetime, logging, argparse,
        np, pd, gpd, xr, plt, ax, fig,
        InferenceSession, SessionOptions,
        open, read, write, load, save, predict, fit,
        transform, fit_transform, run, forward, backward,
        connect, load_collection, save_result, compute_result,
        filter_bbox, filter_temporal, filter_bands,
        filter_spatial, reduce_dimension, apply_dimension,
        aggregate_temporal_period, aggregate_spatial,
        resample_spatial, merge_cubes, ndvi, run_udf,
        mlm_class_random_forest, mlm_class_svm,
        mlm_class_xgboost, mlm_class_catboost,
        mlm_class_mlp, mlm_class_tempcnn,
        mlm_class_tae, mlm_class_lighttae,
        mlm_regr_random_forest, mlm_regr_svm,
        ml_fit, ml_predict, ml_predict_probabilities,
        ml_validate, ml_tune_grid, ml_tune_random,
        ml_uncertainty_class, ml_smooth_class, ml_label_class,
        save_ml_model, load_ml_model, load_stac_ml,
        create_job, start_job, start_and_wait, get_results,
        download_files
    },
    keywordstyle     = [3]\color{lstfunction}\bfseries,
    alsoletter       = {_, .},
    sensitive        = true,
}

\lstdefinelanguage{R}{
    sensitive        = true,
    keywords         = [1]{
        if, else, repeat, while, function, for, in, next, break,
        return, switch, tryCatch, withCallingHandlers, stop, warning,
        message, on.exit, invisible, missing, sys.call, match.arg,
        UseMethod, NextMethod, inherits
    },
    keywords         = [2]{
        TRUE, FALSE, NULL, NA, NaN, Inf, NA_integer_,
        NA_real_, NA_complex_, NA_character_,
        c, list, vector, matrix, array, data.frame, tibble,
        factor, ordered, complex, raw, environment,
        length, dim, nrow, ncol, names, colnames, rownames,
        class, typeof, str, summary, print, cat, paste, paste0,
        sprintf, format, formatC, message, warning, stop,
        library, require, install.packages, source,
        setwd, getwd, dir, file.path, file.exists,
        read.csv, write.csv, readRDS, saveRDS, load, save,
        which, match, seq, seq_len, seq_along,
        rev, sort, order, unique, duplicated, table,
        apply, lapply, sapply, vapply, tapply, mapply, Map,
        Reduce, Filter, Find, Position,
        sum, mean, median, sd, var, min, max, range, cumsum,
        round, floor, ceiling, trunc, abs, sqrt, exp, log,
        sin, cos, tan, is.na, is.null, is.numeric, is.character,
        is.logical, is.list, is.data.frame, is.function,
        as.numeric, as.integer, as.character, as.logical,
        as.data.frame, as.list, as.matrix, as.vector,
        subset, merge, reshape, aggregate, transform,
        plot, ggplot, hist, boxplot, barplot, points, lines,
        par, layout, legend, title, axis, text,
        lm, glm, predict, fitted, residuals, coef, confint,
        anova, aov, t.test, chisq.test, cor, cor.test,
        ts, frequency, start, end, window
    },
    keywords         = [3]{
        connect, login, describe_process, list_collections,
        list_processes, list_file_formats, processes,
        load_collection, save_result, compute_result,
        filter_bbox, filter_temporal, filter_bands,
        filter_spatial, reduce_dimension, apply_dimension,
        aggregate_temporal, aggregate_temporal_period,
        aggregate_spatial, resample_spatial, resample_cube_spatial,
        merge_cubes, rename_labels, rename_dimension,
        ndvi, evi, run_udf, normalized_difference,
        array_element, array_apply, array_filter,
        create_job, start_job,
        gdalcubes, image_collection, cube_view,
        raster_cube, reduce_time, apply_pixel,
        write_tif, write_ncdf, query_gdal,
        rstac, stac, collections, items, stac_search,
        get_request, post_request, assets_download,
        bfastmonitor, bfast,
        sits_tempcnn, sits_tae, sits_lighttae,
        sits_random_forest, sits_svm, sits_xgboost,
        sits_train, sits_classify, sits_accuracy,
        sits_regularize, cube_regularize,
        mlm_class_random_forest, mlm_class_svm,
        mlm_class_xgboost, mlm_class_catboost,
        mlm_class_mlp, mlm_class_tempcnn,
        mlm_class_tae, mlm_class_lighttae,
        mlm_regr_random_forest, mlm_regr_svm,
        ml_fit, ml_predict, ml_predict_probabilities,
        ml_validate, ml_tune_grid, ml_tune_random,
        ml_uncertainty_class, ml_smooth_class, ml_label_class,
        save_ml_model, load_ml_model, load_stac_ml,
        fit_ml_model, predict_ml_model,
        fit_regr_random_forest, fit_class_random_forest,
        predict_random_forest
    },
    morecomment      = [l]{\#},
    morestring       = [b]",
    morestring       = [b]',
    alsoletter       = {., _},
}

\lstdefinestyle{rstyle}{
    style            = basestyle,
    language         = R,
    keywordstyle     = [1]\color{lstkeyword}\bfseries,
    keywordstyle     = [2]\color{lstkeywordB}\bfseries,
    keywordstyle     = [3]\color{lstfunction}\bfseries,
    alsoletter       = {., _},
}

\title{A Machine Learning API for Earth Observation Data Cubes Based on openEO}

\renewcommand{\shorttitle}{A Machine Learning API for EO Data Cubes Based on openEO}
\renewcommand{\headeright}{}
\renewcommand{\undertitle}{}
\date{}

\author{%
  Brian Pondi\textsuperscript{1,*}\,\orcidlink{0009-0008-0367-1690}
  \And
  Jonas Hurst\textsuperscript{1}\,\orcidlink{0009-0000-0004-0051}
  \And
  Rolf Simoes\textsuperscript{2}\,\orcidlink{0000-0003-0953-4132}
  \AND
  Jonas Starke\textsuperscript{1}\,\orcidlink{0009-0000-9749-1923}
  \And
  Marius Appel\textsuperscript{3}\,\orcidlink{0000-0001-5281-3896}
  \And
  Edzer Pebesma\textsuperscript{1}\,\orcidlink{0000-0001-8049-7069}
  \AND
  \begin{minipage}{0.92\textwidth}\centering\normalfont\small
    \textsuperscript{1}Institute for Geoinformatics, University of M\"unster,
    Heisenbergstr.~2, 48149 M\"unster, North-Rhine Westphalia, Germany\\[0.3ex]
    \textsuperscript{2}FGV Agro - Center for Agribusiness Studies, Fundação Getulio Vargas,
    Av.~Paulista, 542, 01310-000 São Paulo, São Paulo, Brazil\\[0.3ex]
    \textsuperscript{3}Bochum University of Applied Sciences,
    Am Hochschulcampus~1, 44801 Bochum, North-Rhine Westphalia, Germany\\[1ex]
    \texttt{brian.pondi@uni-muenster.de}, \texttt{jhurst@uni-muenster.de},
    \texttt{rolfsimoes@gmail.com}, \texttt{jstarke@uni-muenster.de},
    \texttt{marius.appel@hs-bochum.de}, \texttt{edzer.pebesma@uni-muenster.de}\\[0.5ex]
    \textsuperscript{*}Corresponding author: \texttt{brian.pondi@uni-muenster.de}
  \end{minipage}
}

\begin{document}

\maketitle

\begin{abstract}
Earth Observation (EO) data are increasingly organized and analyzed as
spatio-temporal data cubes, while machine learning (ML) methods operate
on tabular feature matrices or structured tensor inputs. This
representational mismatch means that integrating ML into EO workflows currently requires platform-specific transformations that are difficult
to reproduce or transfer across cloud infrastructures. The \mbox{openEO} specification provides a unified interface for EO data access and processing across heterogeneous backends, but lacks a standardized approach for ML integration.

We propose a process-level ML specification for \mbox{openEO} that
structures workflows into three stages: model initialization, model
actions covering training, tuning, inference, validation, and model
management. The specification supports classical algorithms such as
Random Forest and SVM as well as deep learning architectures for
time-series and spatial patch-based modeling, including TempCNN, Temporal Attention Encoders, and foundation model inference. Three
prototype implementations in R and Python demonstrate feasibility
across diverse technology stacks. A crop type mapping use case
demonstrates cross-backend interoperability by submitting an identical
process graph to independent R and Python backends, each training its own model, and comparing the resulting predictions and evaluation metrics. 

Two further use cases demonstrate deep learning on time series and foundation model inference, each executed on a dedicated backend. Prototype
implementations reveal, however, that full cross-backend ML interoperability
requires deeper harmonization of serialization formats and execution
semantics that the process level alone cannot enforce; backend library
versions and preprocessing conventions outside the specification
boundary also affect reproducibility. Explicit backend conformance profiles are the next step to address both. The specification advances the reproducibility, interoperability, and accessibility of ML workflows on EO data cubes across cloud platforms.
\end{abstract}


\section{Introduction}\label{sec1}

Earth Observation (EO) data are increasingly organized and analyzed as
\emph{data cubes}: multidimensional arrays with explicit spatial, temporal,
and band dimensions $(x, y, time, band)$~\cite{baumann2018,appel2019}.
This representation supports scalable analytics, reproducible querying, and
consistent spatio-temporal alignment across large EO data archives.
Driven by advances in satellite technology and open data-sharing policies,
large volumes of EO data are now freely available for applications such as
land-cover monitoring, carbon stock estimation, and disaster
response~\cite{wulder2012,kansakar2016}. Due to the petabyte scale of these
datasets (e.g., more than 90~PB in the Copernicus Data Space
Ecosystem as of May~2026\footnote{\url{https://dashboard.dataspace.copernicus.eu/}, accessed May~2026.}),
EO data cubes are predominantly stored and processed in cloud environments.

\begin{figure}[H]
    \centering
    \includegraphics[width=0.90\textwidth]{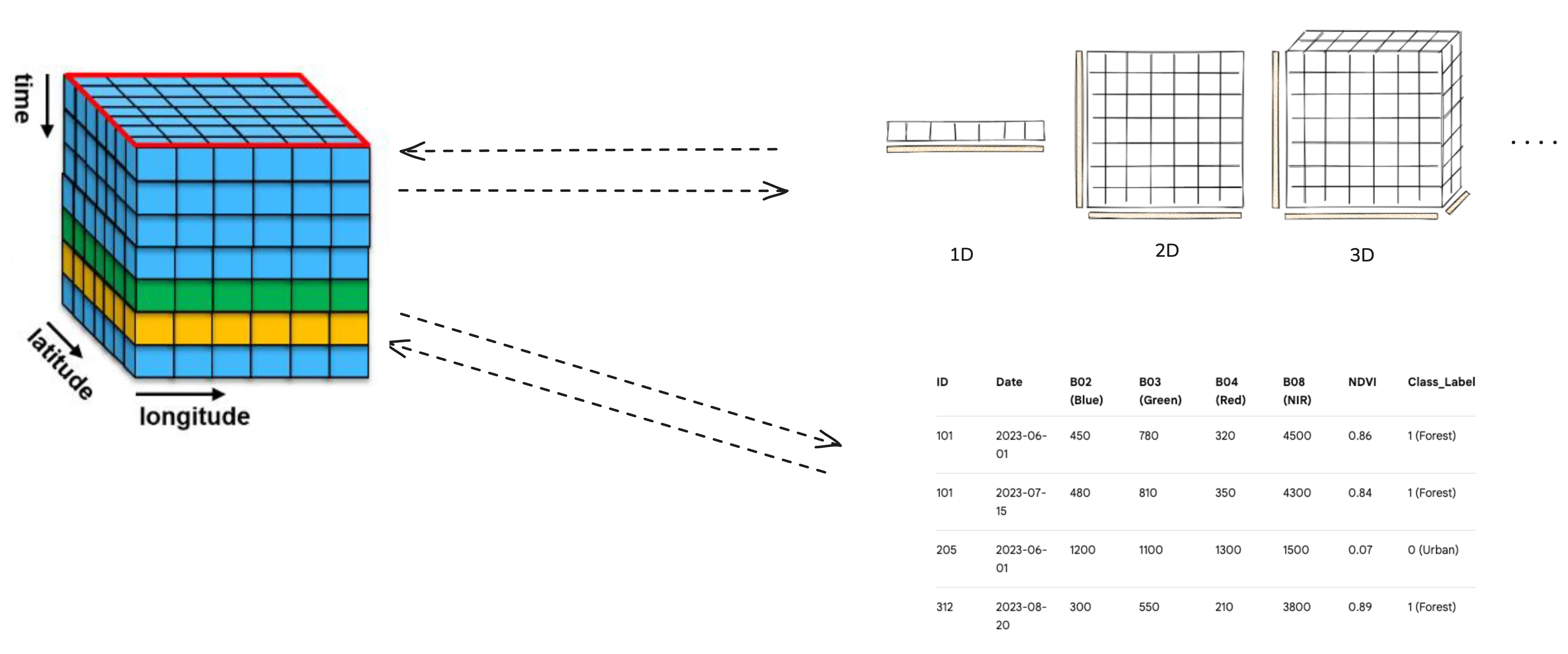}
    \caption{Representation transformations between EO data cubes, tabular
    feature matrices (classical ML), tensor representations (deep learning),
    and prediction cubes. Each transformation must bridge a structural
    mismatch between the cube's named dimensions and the model's expected
    input layout. The feature table and the values in it are illustrative.}
    \label{fig:dc-transform}
\end{figure}

Meanwhile, machine learning (ML) has become a central analytical tool for
extracting patterns and predictions from these datasets, with applications
across land-cover classification~\cite{pelletier2019},
crop yield estimation~\cite{qiao2021}, and disaster
monitoring~\cite{xie2020}. Yet a fundamental representational gap exists
between EO data cubes and what ML models actually consume. Classical methods
such as Random Forests~\cite{breiman2001} and Support Vector
Machines~\cite{cortes1995} expect tabular feature matrices, while deep
learning models require structured multi-dimensional arrays with fixed shape
and ordering. Bridging this gap requires systematic transformations between
cubes, tables, and tensors (Fig.~\ref{fig:dc-transform}) to satisfy each
model's input constraints.

In practice, these transformations are implemented in an \emph{ad hoc} and
platform-specific manner, tightly coupled to particular software stacks or
ML libraries. Workflows developed for one platform are consequently difficult
to reproduce, reuse, or transfer across infrastructures, even when operating
on conceptually identical EO data cubes~\cite{schramm2021}.
Rolf et al.~\cite{rolf2024} have argued that satellite data constitute a
distinct modality in ML whose spatio-temporal structure requires specialized
methods rather than the direct reuse of approaches designed for natural
images. The absence of shared abstractions for EO-specific ML operations
forces each platform to re-implement these methods independently, fragmenting
effort and limiting reproducibility across infrastructures.

The openEO specification standardizes EO data access and processing through
a common API for querying and manipulating data cubes across heterogeneous
cloud backends~\cite{schramm2021}. An openEO analysis is built by
chaining predefined \emph{processes}. Each process performs one well-defined
operation, such as loading a collection, masking clouds, or reducing a
dimension, and is described independently of all others by a published JSON
schema for its parameters and return value. A \emph{process graph} combines
processes into a directed acyclic graph that fixes the order in which they
operate on the data cube. Client libraries build this graph and send it as
JSON to a backend for execution. This makes the process graph the central
data structure exchanged between openEO clients and
backends~\cite{schramm2021}. In April~2026, the Open Geospatial
Consortium (OGC) approved and published the openEO API as an OGC Community
Standard and the openEO Processes as an OGC Community
Practice~\cite{mohr2026openeoapi,mohr2026openeoproc}. As openEO adoption
scales under formal governance, ad~hoc and backend-specific ML
workarounds become increasingly unsustainable: a gap that individual
platforms could manage independently now requires a shared solution at the
specification level. Yet openEO's support for ML remains limited and
fragmented. A small number of ML-related processes exist in isolated
backends and have been labeled as ``experimental'' for more
than four years, notably \texttt{fit\_class\_random\_forest}
and \texttt{predict\_random\_forest} in VITO's Terrascope
platform\footnote{\url{https://docs.terrascope.be/Developers/WebServices/OpenEO/OpenEO.html}}, but these evolved without a shared structure and are not available on other backends. User-defined functions (UDFs) allow deep learning workflows in
principle, but they run in execution environments that each backend
defines and that the openEO specification does not
standardize~\cite{schramm2021,mohr2026openeoapi}. This
limits both interoperability and reproducibility. Extending openEO to support ML,
therefore, requires not just additional endpoints but a coherent
process-level specification that integrates with the existing data cube
model.

To address this gap, we propose a standardized ML specification for EO data
cubes within the \mbox{openEO} ecosystem. The specification structures ML
workflows into three stages: (i)~model initialization, (ii)~model actions
covering training, tuning, inference, and validation, and (iii)~model
management, covering saving and loading. By defining these operations at the
process level, the same workflow can be expressed once and submitted to
any backend that implements them, while remaining agnostic to the underlying ML
representation. We evaluate two claims: (i)~that a process-level ML
specification can express three common supervised ML workflow types
in EO practice (feature-based, time-series, and patch-based) within a
single declarative abstraction, and (ii)~that an identical process graph
submitted to independent \mbox{openEO} backends yields predictions whose
overall accuracy and dominant per-location class assignment agree within
the variation expected from differences in underlying ML library defaults.
Each backend fits its own model from the same inputs; moving a trained
model between backends is not part of this evaluation.

The contributions of this work are fourfold: (i)~a backend-agnostic,
process-level ML specification integrated with the openEO process catalog,
covering classical ML, deep learning on time series, and foundation-model
workflows; (ii)~prototype implementations in both R and Python;
(iii)~representative use cases that show what the specification achieves
for interoperability, reproducibility, and accessibility of ML workflows on
EO data cubes, and where its current limits lie; and (iv)~a proposed L3-ML
capability profile through which backends can declare their ML support in
machine-readable form. The modular process structure provides a foundation for future
extension to emerging paradigms, including federated
learning~\cite{morenoalvarez2024} and large-scale foundation
models~\cite{xiao2024}.

\section{Background}\label{sec2}

\subsection{Data Cubes}

A widely used data structure in EO is the \emph{raster data cube}: a
multidimensional array with consistent spatial, temporal, and spectral axes
that organizes large satellite image collections into a uniform, queryable
form~\citep{kopp2019}. Image collections comprise sets of satellite images,
each containing multiple spectral bands or variables. Within a single image,
bands typically share a common spatial extent, acquisition timestamp, and
reference system, although their pixel resolutions may differ~\citep{appel2019}.
Unlike the regular structure of data cubes, conventional image collections
often include scenes with irregular spatial coverage, variable acquisition
intervals, and differing coordinate reference systems, making uniform
processing more challenging.

By standardizing indexing along named dimensions $(x, y, time, band)$ and
attaching explicit metadata, data cubes are designed so that the same
sequence of operations yields the same values on different backends, provided those
backends apply the same resampling, compositing, and masking rules.
Deterministic re-slicing and consistent dimension labels mean that training, validation, and inference
draw on the same spatio-temporal references regardless of the execution
environment. These properties make the data cube the natural anchor for the
ML specification proposed in this paper: the initialization, fit, predict,
and management processes defined in later sections all operate on this shared
structure.

\begin{figure}[H]
    \centering
    \includegraphics[width=0.8\textwidth]{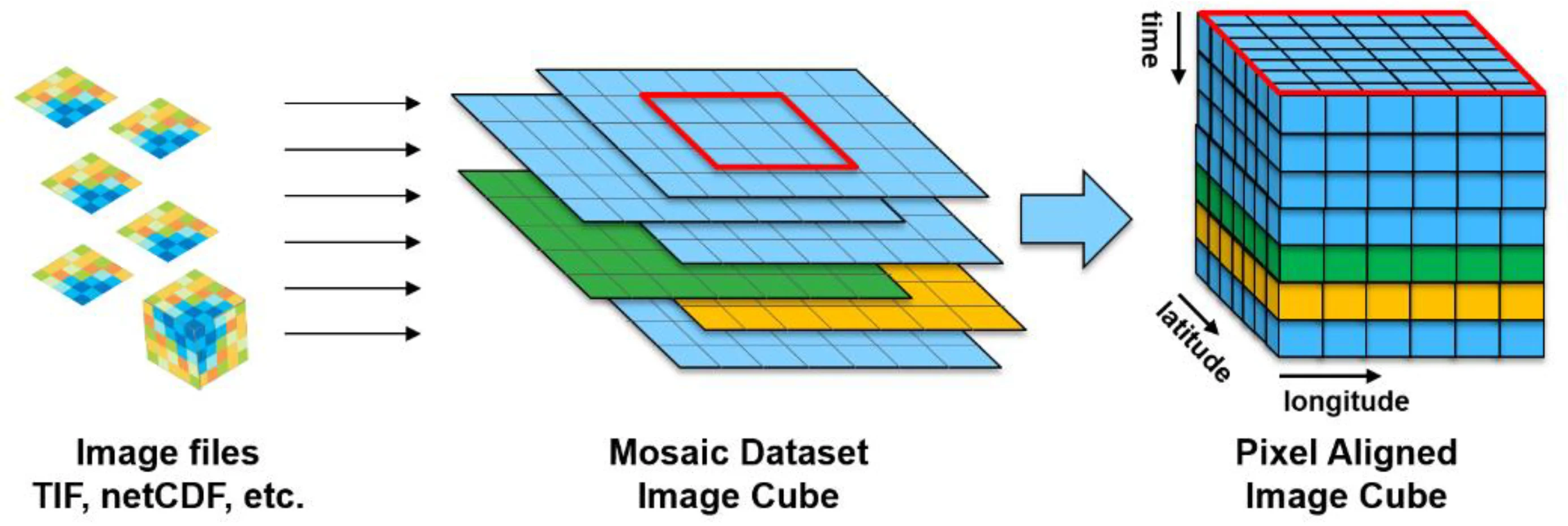}
    \caption{A workflow for converting image/raster files into a
    pixel-aligned image/raster data cube $(x, y, time, band)$.
    Adapted from~\cite{kopp2019}.}
    \label{fig:dc}
\end{figure}

Figure~\ref{fig:dc} illustrates a workflow for converting image files into
the canonical cube layout $(x, y, time, band)$ that underpins sampling and
batching throughout the specification.

The regular, tensor-like structure of raster data cubes makes them naturally
compatible with a wide range of ML architectures. Convolutional and recurrent
models benefit directly from the spatial and temporal ordering that cubes
preserve~\cite{zhu2017}, and architectures developed specifically for EO
data, including \mbox{TempCNN}~\cite{pelletier2019},
\mbox{ConvLSTM}~\cite{russwurm2018}, and Temporal Self-Attention
networks~\cite{saintefaregarnot2020}, exploit this structure to learn
phenological and spatial patterns from aligned observation sequences.

Complementing raster cubes, \emph{vector data
cubes}\footnote{\url{https://r-spatial.org/r/2022/09/12/vdc.html}} provide a
structured representation for training and validation data. Like raster cubes,
they are modeled as $n$-dimensional arrays, but include at least one spatial
dimension that maps to a set of 2-D vector geometries such as points, lines,
or polygons. Additional axes capture time, attributes, or category labels.
This design enables vector cubes to store ground-truth annotations such as
field parcel boundaries, land-cover classes, or point measurements in a
queryable form. Vector data cubes can also be joined with raster data
cubes by spatial and temporal relationships, providing a clean interface
between feature inputs and target variables in supervised learning workflows.

\subsection{ML Workflow Types for EO Data Cubes}\label{sec:workflow_types}

EO data present distinctive challenges for ML. Unlike conventional tabular
datasets, satellite imagery is inherently spatio-temporal, high-dimensional,
and spatially autocorrelated~\cite{rolf2024,adegun2023}. Observations combine
multiple spectral bands, fine spatial resolution, and repeated temporal
acquisitions, characteristics that violate the independent and identically
distributed (i.i.d.) assumption underlying many standard ML methods.
Linear regression, logistic regression, and naive Bayes classifiers assume
that observations are drawn independently and identically from a fixed
distribution, an assumption that breaks down when adjacent pixels share
spatial context or when repeated acquisitions are temporally correlated.
Even ensemble methods such as Random Forests~\cite{breiman2001} and gradient
boosting machines, while more robust in practice, rely on tabular feature
vectors that must be carefully engineered to encode spatio-temporal
structure explicitly. Support Vector Machines~\cite{cortes1995} face similar
constraints. ML workflows for EO must therefore explicitly account for
spatial and temporal structure rather than treating observations as
exchangeable samples.

Within openEO, satellite collections are exposed as data cubes that can be
queried and transformed through standardized processes. This abstraction
enables consistent extraction of pixel time series, image patches, or
aggregated features across backends. However, different ML models consume EO
data in fundamentally different forms, and the way a data cube is decomposed
into learning units shapes the entire ML workflow and determines which
operations must be supported. The three use cases in
Sections~\ref{sec:use-case-1}, \ref{sec:use-case-2},
and~\ref{sec:use-case-3} test whether a process-level abstraction can
express each of the three decomposition strategies described below.

Three ways of turning a data cube into model input recur in the remote
sensing literature: engineered feature vectors for classical
classifiers~\cite{belgiu2016,maxwell2018}, pixel or parcel time series for
sequence models~\cite{pelletier2019,miller2024}, and image patches for
convolutional and attention-based networks~\cite{zhu2017,ma2019}. We refer
to these as three workflow types, each
defined by how the data cube is transformed into the input structure
expected by the model.

\textbf{Feature-based workflows} convert observations at each location into
engineered descriptors, including spectral indices, textural measures, and
ancillary layers, aggregated into a two-dimensional tabular feature matrix
$(N \times F)$, where $N$ is the number of samples and $F$ the number of
features. This matrix is the canonical input for classical estimators such
as Random Forests or SVMs~\cite{belgiu2016,maxwell2018}. Supporting this
workflow type requires processes for feature extraction, spatial
aggregation, and sampling that transform a cube into this canonical form.

\textbf{Time-series workflows} treat each cube location as a temporal
signal, modeling phenological trajectories or disturbance dynamics directly
from observation sequences rather than engineered
features~\cite{simoes2021}. Deep sequence models such as
TempCNNs~\cite{pelletier2019} and attention-based
architectures~\cite{saintefaregarnot2020} consume inputs shaped as
$(B \times T \times C)$, where $B$ is batch size, $T$ the number of time
steps, and $C$ the number of spectral channels. Supporting this workflow
type requires processes for temporal interpolation, gap-filling, and
consistent extraction of ordered sequences from data cubes.

\textbf{Spatial patch-based workflows} leverage local spatial context by
operating on patches extracted from data cubes, enabling convolutional
networks to capture texture, structure, and spatial
dependencies~\cite{zhu2017,russwurm2018,qiao2021}. Two-dimensional patch
models consume inputs shaped as $(B \times C \times H \times W)$, while
spatio-temporal models extend this to $(B \times T \times C \times H \times
W)$~\cite{shi2015} or full 4-D stacks $(B \times T \times C \times Z \times
Y \times X)$ for volumetric representations~\cite{bodnar2024}, where $H$,
$W$, and $Z$ are the spatial and vertical extents of the patch. Bi-temporal change
detection models consume a specialised variant shaped as $(B \times 2 \times
C \times H \times W)$, where the second dimension indexes the two acquisition
dates being compared~\cite{rolih2025}. Supporting this workflow type requires
tiling, patch extraction, batching, and reconstruction of predictions into
spatially coherent output cubes.

In operational settings these workflow types frequently intersect: feature
engineering may precede temporal modeling, and hybrid architectures often
combine spatial and temporal context. The three types were selected because
they cover widely used supervised ML paradigms in EO
practice and represent the breadth of operations a general specification
must support. Emerging approaches such as graph-based learning or large
foundation models could be incorporated incrementally through the same
process-level abstraction, although this is not tested here.

\subsection{ML API Conventions and Standards}
\label{sec:ml-background}
 
Covering this range of workflow types within a single specification
requires grounding in both EO-specific requirements and the API
conventions that have emerged in the broader ML ecosystem.
Supervised ML APIs have converged on a design pattern that
scikit-learn~\cite{pedregosa2011} established: configure a model with
hyperparameters, train it via \texttt{fit(X,\,y)}, and apply it via
\texttt{predict(X)}, with each stage kept separate. PyTorch's
\texttt{Module}\footnote{\url{https://docs.pytorch.org/docs/2.12/notes/modules.html}} abstraction and the Keras \texttt{compile}/\texttt{fit}
API\footnote{\url{https://keras.io/api/models/model_training_apis/}} carry the same separation into deep learning; the Hugging Face
\texttt{Transformers} library~\cite{wolf2020} extends it to pretrained Natural Language model pipelines, and MLflow~\cite{chen2020} applies it to model lifecycle management through experiment tracking and model registries. The proposed specification's three-stage structure, introduced in Section~\ref{sec3}, maps directly onto this same separation, transposed to the declarative process-level abstraction of \mbox{openEO} and the dimensional structure of EO data cubes.

ONNX (Open Neural Network Exchange)~\cite{onnx}, an open standard
governed by the Linux Foundation and supported by a consortium of
hardware and software vendors, represents trained ML models as
computation graphs independent of the training framework. A
model trained in PyTorch, scikit-learn, Tensorflow, XGBoost, LibSVM, or Keras can be exported to ONNX and run in ONNX-compatible runtimes that support the operators it uses, separating training from deployment. Because EO backends may rely on different ML execution stacks, as the prototype implementations in Section~\ref{sec4} illustrate, models need an exchange format. ONNX is one option, and currently the only widely supported one that is independent of the training framework. In principle, a model produced on one backend can then be loaded and applied on another without bridging code. In practice, conversion is a source of error in its own right: a survey and failure analysis of ONNX converters found that about a third of the reported failures concerned semantically incorrect models, that is, converted models whose behaviour differed from the original~\cite{jajal2024}, and an evaluation of EO deep learning models across Python, R, and Julia found ONNX portability to be architecture dependent in ways that export success alone does not predict~\cite{pondi2026onnx}. An ONNX export therefore has to be checked numerically against the source model before it is trusted on a second backend. Framework-specific formats remain valid when producer and consumer share a framework, so the specification does not prescribe a format but records it in the model metadata (Section~\ref{sec:model-management}).

The SpatioTemporal Asset Catalog
(STAC)\footnote{\url{https://stacspec.org}} is an open JSON
specification for cataloguing geospatial data assets with queryable,
provider-agnostic metadata, used by major EO data providers and open
data archives. The STAC Machine Learning Model (STAC~MLM)
extension~\cite{charette2024} applies this framework to spatio-temporal ML models, standardizing how architecture, input and output specifications, model hyperparameters, pretrained source, tasks and software dependencies are recorded as catalog assets. Because openEO backends already interact with STAC-compliant catalogs for data access, aligning model management with the same infrastructure means trained models can be shared and retrieved through workflows EO practitioners already have in place.

\section{ML API Specification}\label{sec3}
Building on the workflow types in Section~\ref{sec:workflow_types}
and the API conventions in Section~\ref{sec:ml-background}, this
section describes the ML API processes for the \mbox{openEO} ecosystem.
The specification is organized into three stages: model initialization,
model actions, and model management (Sections~\ref{sec:model-init} to~\ref{sec:model-management}), each covering a distinct subset of
processes for end-to-end ML on EO data cubes. Section~\ref{sec:capabilities} then describes how a backend
declares which of these processes, model formats, and workflow types it
supports.

The proposed processes are described as JSON schemas, available in the
project repository on
GitHub\footnote{\url{https://github.com/PondiB/openEO-processes}}.
An example schema is provided in Appendix~\ref{secB1}. This follows the
openEO community's standard development workflow: GitHub is the platform
for proposing and iterating on process specifications through open issues
and pull requests. Once a specification has been implemented in at least
two independent production backends, it is promoted to the stable,
versioned catalogue\footnote{\url{https://processes.openeo.org/}}, which is the
normative specification. The broader openEO Processes
specification, which defines the framework within which the proposed ML
processes sit, has been approved and published as an OGC Community
Practice (version~1.2, April 2026)\footnote{\url{https://docs.ogc.org/cp/24-060.html}}, situating this work within
an established international open geospatial standards context.

Five core principles defined by the authors guided the design of the API:

\begin{itemize}
    \item \textbf{Modularity}: Decoupling initialization, action, and
    management into discrete processes allows users to compose complex
    pipelines from simple building blocks and swap algorithms without
    rewriting entire workflows.

    \item \textbf{Consistency}: Uniform naming conventions, parameter
    ordering, and behavior across all ML processes minimize surprises for
    users, improving learnability and reducing integration friction.

    \item \textbf{Backend-Agnosticism}: Defining processes independently of
    execution engines allows backends to map ML processes to native libraries
    such as scikit-learn, XGBoost, PyTorch, or caret, while preserving a
    consistent user experience.

    \item \textbf{Sensible Defaults}: Processes expose well-chosen default
    hyperparameters that cover the majority of common use cases, so that
    simple ML tasks run without configuration, reducing errors and
    accelerating prototyping.

    \item \textbf{Extensibility}: A compact core API covers common ML tasks
    while supporting extension through community proposals, enabling new
    algorithms and workflow types to be added incrementally without
    modifying existing processes.
\end{itemize}

The selection of processes presented in this section is intentionally
pragmatic rather than exhaustive, and the rationale for inclusion is
explicit. Processes were selected on three criteria: (i)~they correspond to
ML capabilities already widely used in operational EO analysis; (ii)~they
cover at least one of the three workflow types identified in
Section~\ref{sec2}, ensuring that feature-based, time-series, and spatial
patch-based workflows are all supported; and (iii)~mature, tested
implementations exist in established libraries so that backends can realize
the processes without bespoke ML development. Processes that do not yet
meet criterion (iii), such as graph neural network architectures, are
deliberately excluded from the current specification and left for future
community proposals. In particular, several classification processes were
introduced to make methods available in EO-focused libraries, such as the
\texttt{sits} package~\cite{simoes2021} in R, accessible within the
\mbox{openEO} ecosystem. This scoped selection provides a concrete starting
point while keeping the process set open for future extension.

\subsection{Model Initialization}\label{sec:model-init}

Model initialization processes carry the prefix \texttt{mlm\_} and produce
untrained model definition objects. Each process name follows the pattern:
\begin{center}
  \texttt{mlm\_<type>\_<model>}
\end{center}
where \texttt{<type>} is an abbreviated ML task category, such as
\texttt{class} for classification, \texttt{regr} for regression,
\texttt{segm} for segmentation, and \texttt{gen} for generative models, and
\texttt{<model>} identifies the algorithm or architecture, such as
\texttt{random\_forest}, \texttt{svm}, \texttt{xgboost}, \texttt{tempcnn},
or \texttt{tae}.

Each initialization process returns a \texttt{Model} object that captures
the selected algorithm together with its initial hyperparameters and
references to the corresponding training and inference routines. This
representation allows users to define models ranging from classical
algorithms to deep learning architectures, while deferring all computation
to later process stages.

All initialization processes expose a \texttt{seed} parameter for
controlling stochastic behavior and supporting reproducible experimentation.
Hyperparameters are given sensible default values so that models can be
trained immediately without configuration; any parameter can be
explicitly overridden during the initialization call to suit the
specific task. These parameters can alternatively
be designated as tuning candidates for the dedicated search processes
\texttt{ml\_tune\_grid}, which evaluates a user-defined grid of
configurations, and \texttt{ml\_tune\_random}, which samples configurations
from user-defined distributions~\cite{bergstra2012}.

\textbf{Integrating new algorithms.} The specification is designed so that
new model types can be added without modifying any existing process. A
contributor adds a new \texttt{mlm\_} process by: (i)~choosing a name that
follows the \texttt{mlm\_<type>\_<model>} convention; (ii)~defining the
process schema in JSON, specifying parameters, their types, and default
values following the same structure shown in Appendix~\ref{secB1};
(iii)~returning a \texttt{Model} object whose \texttt{train} field holds the
backend training routine and whose \texttt{predict} field holds the
inference routine. Once registered, the new process is immediately
accessible through \texttt{ml\_fit} and \texttt{ml\_predict} without any
changes to those processes. The \mbox{openEOcraft} decorator mechanism
illustrated in Section~\ref{sec:use-case-2} demonstrates this pattern
concretely: a single decorated R function becomes a fully registered
\mbox{openEO} process accessible to all client libraries.
Table~\ref{tab:ml-init} lists the representative initialization processes
currently included in the specification.

\begin{table}[!htbp]
    \centering
    \caption{Examples of ML Initialization Processes (mlm\_ prefix) with
    representative tunable parameters.}
    \label{tab:ml-init}
    \begin{tabular}{@{\hspace{0.5em}}p{0.28\textwidth}@{\hspace{0.04\textwidth}}p{0.63\textwidth}@{\hspace{0.5em}}}
        \toprule[1.2pt]
        \textbf{Process Name} & \textbf{Description} \\
        \midrule[1pt]

        \texttt{mlm\_class\_catboost} &
        Initializes a CatBoost classification model for later training using
        \texttt{ml\_fit}. Typical tunable parameters include the number of
        trees, learning rate, tree depth, loss function, and random seed. \\
        \addlinespace[0.5em]

        \texttt{mlm\_class\_mlp} &
        Initializes a Multi-Layer Perceptron classification model. Tunable
        parameters include the number of hidden layers, units per layer,
        activation functions, learning rate, batch size, and training
        epochs. \\
        \addlinespace[0.5em]

        \texttt{mlm\_class\_random\_forest} &
        Initializes a Random Forest classification model for later training
        using \texttt{ml\_fit}. Example tunable parameters include the
        number of trees and the maximum number of variables per split. \\
        \addlinespace[0.5em]

        \texttt{mlm\_class\_svm} &
        Initializes a Support Vector Machine classification model. Tunable
        parameters include the kernel type, regularization parameter, and
        kernel coefficient. \\
        \addlinespace[0.5em]

        \texttt{mlm\_class\_xgboost} &
        Initializes an XGBoost classification model for later training using
        \texttt{ml\_fit}. Representative tunable parameters include learning
        rate, maximum tree depth, subsampling ratio, and regularization
        terms. \\
        \addlinespace[0.5em]

        \texttt{mlm\_class\_tempcnn} &
        Initializes a Temporal Convolutional Neural Network model for time
        series classification. Tunable parameters include the number of
        convolutional layers, kernel sizes, dropout rates, optimizer
        settings, and learning rate. \\
        \addlinespace[0.5em]

        \texttt{mlm\_class\_tae} &
        Initializes a Temporal Attention Encoder model for time series
        classification. Example tunable parameters include epochs, weight
        decay, batch size, optimizer, and learning rate. \\
        \addlinespace[0.5em]

        \texttt{mlm\_class\_lighttae} &
        Initializes a lightweight Temporal Attention Encoder model. Tunable
        parameters typically include epochs, weight decay, batch size,
        optimizer, and learning rate. \\
        \addlinespace[0.5em]

        \texttt{mlm\_regr\_svm} &
        Initializes a Support Vector Machine regression model. Tunable
        parameters include kernel type, regularization strength, kernel
        coefficient, and the epsilon-insensitive loss parameter. \\
        \addlinespace[0.5em]

        \texttt{mlm\_regr\_random\_forest} &
        Initializes a Random Forest regression model for later training using
        \texttt{ml\_fit}. Example tunable parameters include the number of
        trees and the maximum number of variables per split. \\

        \bottomrule[1.2pt]
    \end{tabular}
\end{table}

\subsection{Model Actions}\label{sec:model-actions}

Model action processes carry the prefix \texttt{ml\_} and execute the core
stages of an ML workflow: training, hyperparameter tuning, validation,
inference, uncertainty estimation, and spatial post-processing. Each process
name follows the pattern:
\begin{center}
  \texttt{ml\_<action>}
\end{center}

The central process in this group is \texttt{ml\_fit}, which takes the
untrained \texttt{Model} object produced by any initialization process
and trains it against a vector data cube containing input features and
target labels. In this iteration of the specification, train and test
partitioning is handled internally; users provide their full labelled
dataset without pre-splitting. Once trained, the model is passed to
\texttt{ml\_predict} for predictions, or to
\texttt{ml\_predict\_probabilities} for probability outputs on classification tasks. Both prediction processes also accept models loaded from serialized
formats (Section~\ref{sec:model-management}).

These probability outputs feed directly into downstream processes.
\texttt{ml\_uncertainty\_class} quantifies prediction confidence using
measures such as least-confidence, margin, or probability ratio, enabling
human-in-the-loop workflows by highlighting regions where model certainty is
low or training data are sparse. \texttt{ml\_smooth\_class} applies Bayesian
spatial smoothing to probability outputs, reducing per-pixel classification
noise and improving spatial coherence in scenes with mixed land cover
types~\cite{simoes2021}. Finally, \texttt{ml\_label\_class} converts a
probability cube into a labeled output cube by applying softmax normalization
and selecting the class with the highest probability at each location.

Hyperparameter search is supported through \texttt{ml\_tune\_grid} and
\texttt{ml\_tune\_random}, and model evaluation through
\texttt{ml\_validate}, which computes user-selected performance metrics
against a user-provided validation set, and \texttt{ml\_validate\_kfold},
which accepts an untrained model and a full training set, manages fold
assignment internally, and returns the model refitted on the full
training set alongside cross-validation scores.
Table~\ref{tab:ml-actions} summarizes the full set of model action
processes. The design is open for extension, and new processes can be
contributed under the same naming convention.

\begin{table}[h!]
    \centering
    \caption{ML Action Processes (ml\_ prefix), including training, tuning,
    validation, and inference.}
    \label{tab:ml-actions}
    \begin{tabular}{@{\hspace{0.5em}}p{0.28\textwidth}@{\hspace{0.04\textwidth}}p{0.63\textwidth}@{\hspace{0.5em}}}
        \toprule[1.2pt]
        \textbf{Process Name} & \textbf{Description} \\
        \midrule[1pt]

        \texttt{ml\_fit} &
        Fits a machine learning model to a data cube of input features and
        target values. Trains the model using the provided data and returns
        a trained model object. \\
        \addlinespace[0.5em]

        \texttt{ml\_tune\_grid} &
        Performs grid-based hyperparameter tuning for an initialized machine
        learning model using user-defined parameter ranges and evaluation
        criteria. \\
        \addlinespace[0.5em]

        \texttt{ml\_tune\_random} &
        Performs randomized hyperparameter tuning for an initialized machine
        learning model by sampling parameter configurations from user-defined
        distributions. \\
        \addlinespace[0.5em]

        \texttt{ml\_validate} &
        Evaluates a trained machine learning model against a user-provided
        validation set and computes user-selected performance metrics. \\
        \addlinespace[0.5em]

        \texttt{ml\_validate\_kfold} &
        Accepts an untrained model and a full training set, performs
        $k$-fold cross-validation with internally managed fold assignment,
        and returns the model refitted on the full training set enriched
        with aggregate per-class and overall cross-validation scores. \\
        \addlinespace[0.5em]

        \texttt{ml\_predict} &
        Applies a trained machine learning model to a data cube of input
        features and returns the predicted values. \\
        \addlinespace[0.5em]

        \texttt{ml\_predict\_probabilities} &
        Applies a model to input features and returns predicted class
        probabilities. Adds a new classes dimension with probability values
        for each class. \\
        \addlinespace[0.5em]

        \texttt{ml\_uncertainty\_class} &
        Estimates classification uncertainty using methods such as margin,
        ratio, or least-confidence based on predicted probability
        distributions. \\
        \addlinespace[0.5em]

        \texttt{ml\_smooth\_class} &
        Applies spatial smoothing to classification probability results using
        Bayesian inference, improving spatial coherence in heterogeneous
        scenes. \\
        \addlinespace[0.5em]

        \texttt{ml\_label\_class} &
        Converts a probability data cube to a labeled data cube by applying
        softmax normalization and selecting the class with the highest
        probability. \\

        \bottomrule[1.2pt]
    \end{tabular}
\end{table}

All processes that train a model, namely \texttt{ml\_fit},
\texttt{ml\_tune\_grid}, \texttt{ml\_tune\_random}, and
\texttt{ml\_validate\_kfold}, take their \texttt{training\_set} as a vector
data cube. It has a \texttt{geometry} dimension with one point or polygon per
labelled sample, in a declared coordinate reference system, and one
\texttt{bands} (or \texttt{other}) dimension whose labels name the
predictors. Time-series models additionally expect an ordered temporal
dimension, so that each sample carries a $(T \times C)$ array of
observations. The \texttt{target} parameter names the variable that holds
the label: a string or integer class for classification, or a number for
regression. This is exactly the structure that \texttt{aggregate\_spatial}
produces from a raster cube and a set of labelled geometries, so a training set can be built entirely from
standard openEO processes and an open vector format such as GeoJSON.

The schema also accepts a string, which lets a user reference an existing,
published training set by URL instead of rebuilding it inside the process
graph. How such a file is parsed is currently left to the backend.
Package-specific formats, such as the \texttt{sits} samples used in
Section~\ref{sec:use-case-2}, are backend extensions rather than part of the
specification: a backend that accepts them must say so in its capability profile
(Section~\ref{sec:capabilities}), and the same samples can always be
supplied as a vector data cube.

\subsection{Model Management}\label{sec:model-management}

Model management processes handle the storage and retrieval of ML artifacts,
enabling models to be saved, shared, discovered, and reloaded across sessions
and backends. The processes follow existing \mbox{openEO} naming conventions,
as listed in Table~\ref{tab:data-management}.

Saving and loading are aligned with the STAC~MLM
extension~\cite{charette2024} (see Section~\ref{sec:ml-background}),
so that saved models are catalogued as reproducible assets
discoverable by other users and backends. For models that are intended for
local use rather than broad sharing, the \texttt{load\_ml\_model} process
provides direct access to models already registered within the current
backend. The openEO clients can later on find and list these models through the proposed openEO API level endpoint \texttt{GET /ml\_models}, which lists them with their STAC MLM metadata, as
\texttt{GET /collections} does for \texttt{load\_collection}
(Section~\ref{sec:capabilities}).

\begin{table}[h!]
    \centering
    \caption{Model Management Processes.}
    \label{tab:data-management}
    \begin{tabular}{@{\hspace{0.5em}}p{0.28\textwidth}@{\hspace{0.04\textwidth}}p{0.63\textwidth}@{\hspace{0.5em}}}
        \toprule[1.2pt]
        \textbf{Process Name} & \textbf{Description} \\
        \midrule[1pt]

        \texttt{load\_ml\_model} &
        Loads a machine learning model from the current backend by its ID
        and returns it for use in subsequent processes. \\
        \addlinespace[0.5em]

        \texttt{load\_stac\_ml} &
        Loads a machine learning model from a STAC MLM Item into the current
        session, enabling reuse of externally catalogued models. \\
        \addlinespace[0.5em]

        \texttt{save\_ml\_model} &
        Saves a trained machine learning model and produces an accompanying
        STAC Item compliant with the STAC MLM Extension, making the model
        discoverable and reusable by others. \\

        \bottomrule[1.2pt]
    \end{tabular}
\end{table}

The STAC Item written by
\texttt{save\_ml\_model} and read by \texttt{load\_stac\_ml} follows version
1.4 or later of the STAC MLM extension\footnote{\url{https://github.com/stac-extensions/mlm}}.
Table~\ref{tab:mlm-metadata} lists the fields the specification relies on.
The fields marked as required by STAC MLM are mandatory for every saved
model. The remaining fields are optional in STAC MLM, but
\texttt{save\_ml\_model} must fill them whenever the backend knows their
value, because \texttt{load\_stac\_ml} needs them to decide whether a model
can run at all. Most values do not have to be supplied by the user: the
backend derives them from the \texttt{mlm\_} initialization process, the
training set, and its own software environment.

\begin{table}[h!]
    \centering
    \caption{STAC fields used by \texttt{save\_ml\_model} and
    \texttt{load\_stac\_ml}. R: required by STAC MLM (v1.4 or later); F: filled by
    the backend whenever known. Names in \texttt{dim\_order} are the dimension
    names of the data cube, which is how \texttt{ml\_predict} maps cube
    dimensions to model dimensions.}
    \label{tab:mlm-metadata}
    \small
    \begin{tabular}{@{\hspace{0.3em}}>{\raggedright\arraybackslash}p{0.34\textwidth}@{\hspace{0.02\textwidth}}c@{\hspace{0.02\textwidth}}>{\raggedright\arraybackslash}p{0.52\textwidth}@{\hspace{0.3em}}}
        \toprule[1.2pt]
        \textbf{Field} & \textbf{Status} & \textbf{Source when saving} \\
        \midrule[1pt]
        \texttt{mlm:name} & R & model name \slash   \: \texttt{name} argument of \texttt{save\_ml\_model} \\
        \texttt{mlm:architecture} & R & Algorithm or architecture given by \texttt{<model>} in the initialization process name (e.g.\ Random Forest, TempCNN) \\
        \texttt{mlm:tasks} & R & Task given by \texttt{<type>} in the initialization process name, mapped to the STAC MLM task vocabulary (e.g.\ \texttt{regression}; \texttt{classification} for per-pixel classes) \\
        \addlinespace[0.3em]
        \texttt{mlm:input}: \texttt{name}, \texttt{input} (\texttt{shape}, \texttt{dim\_order}, \texttt{data\_type}) & R & Structure of the training data cube as passed to the model: samples become the \texttt{batch} dimension ($-1$), followed by \texttt{bands} and, if present, \texttt{time}, \texttt{y}, \texttt{x}; data type of the model input \\
        \quad\texttt{bands} (or \texttt{variables}) & F & Labels of the band dimension in model order; the Item must also define these bands (or data cube variables) \\
        \quad\texttt{value\_scaling} & F & Normalization applied to the features during training \\
        \addlinespace[0.3em]
        \texttt{mlm:output}: \texttt{name}, \texttt{tasks}, \texttt{result} (\texttt{shape}, \texttt{dim\_order}, \texttt{data\_type}) & R & Structure of the raw model output, e.g.\ \texttt{batch} for one label per sample or \texttt{batch}, \texttt{class} for class probabilities \\
        \quad\texttt{classification:classes} & F & Distinct values of the target variable (classification only) \\
        \addlinespace[0.3em]
        Asset with role \texttt{mlm:model} and \texttt{mlm:artifact\_type} & R & Serialization used by the backend (e.g.\ \texttt{torch.jit.save}, \texttt{pickle.dump}, ONNX, R object) \\
        \texttt{mlm:framework}, \texttt{mlm:framework\_version} & F & Backend software environment \\
        \texttt{mlm:hyperparameters} & F & Initialization arguments, including \texttt{seed} \\
        \texttt{mlm:accelerator}, \texttt{mlm:accelerator\_constrained} & F & Hardware intended for inference (e.g.\ \texttt{amd64}, \texttt{cuda}) \\
        \texttt{file:checksum} (STAC File extension) & F & Computed from the stored artifact \\
        \bottomrule[1.2pt]
    \end{tabular}
\end{table}

The metadata may describe a model that the
executing backend cannot run. The specification therefore requires
backends to fail early and explicitly rather than adapt silently. When
\texttt{load\_stac\_ml} or \texttt{load\_ml\_model} is evaluated, the backend
rejects the model if it does not support the declared
\texttt{mlm:framework} or \texttt{mlm:artifact\_type}, if the installed
framework differs from \texttt{mlm:framework\_version} in its major version,
or if \texttt{mlm:accelerator\_constrained} is true and the declared
accelerator is not available. A difference in minor version is allowed but
must be reported in the job log. Before inference, \texttt{ml\_predict}
rejects the data cube if a dimension named in \texttt{mlm:input} is missing,
if a dimension is shorter than the model input, or if a required band is
absent and cannot be computed. If the tensor returned by the model does not
match the shape or data type declared in \texttt{mlm:output}, the job fails
instead of reshaping the result. In every case the error names the
offending field, so the mismatch can be traced to either the metadata or
the runtime.

\subsection{Capability Discovery}\label{sec:capabilities}

A client, or a federation broker that forwards jobs to member
backends~\cite{mohr2025}, has to know before submission whether a backend
can train a given model type, load an ONNX file, or run patch-based
inference. \texttt{GET /processes} lists available processes but not the
frameworks, model formats, or workflow types behind them, and the existing
openEO L3-ML process
profile\footnote{\url{https://openeo.org/documentation/1.0/developers/profiles/processes.html}}
lists only a few legacy ML processes. Appendix~\ref{secD1} therefore proposes
an extended L3-ML process profile and a companion API profile with two
endpoints at the level of the openEO API: \texttt{GET /ml\_runtimes}, which
declares a backend's ML frameworks, model formats, and workflow types, and
\texttt{GET /ml\_models}, which lists the models stored on the backend.
\mbox{openEOcubes} and \mbox{openEOcraft} already serve
\texttt{GET /ml\_runtimes} and \texttt{GET /ml\_models}.
\mbox{openeo-processes-dask-ml} is a process library rather than a full
backend and depends on the backend that hosts it.

\section{Prototype Implementations}\label{sec4}

\begin{figure}[!hbt]
    \centering
    \includegraphics[width=0.8\textwidth]{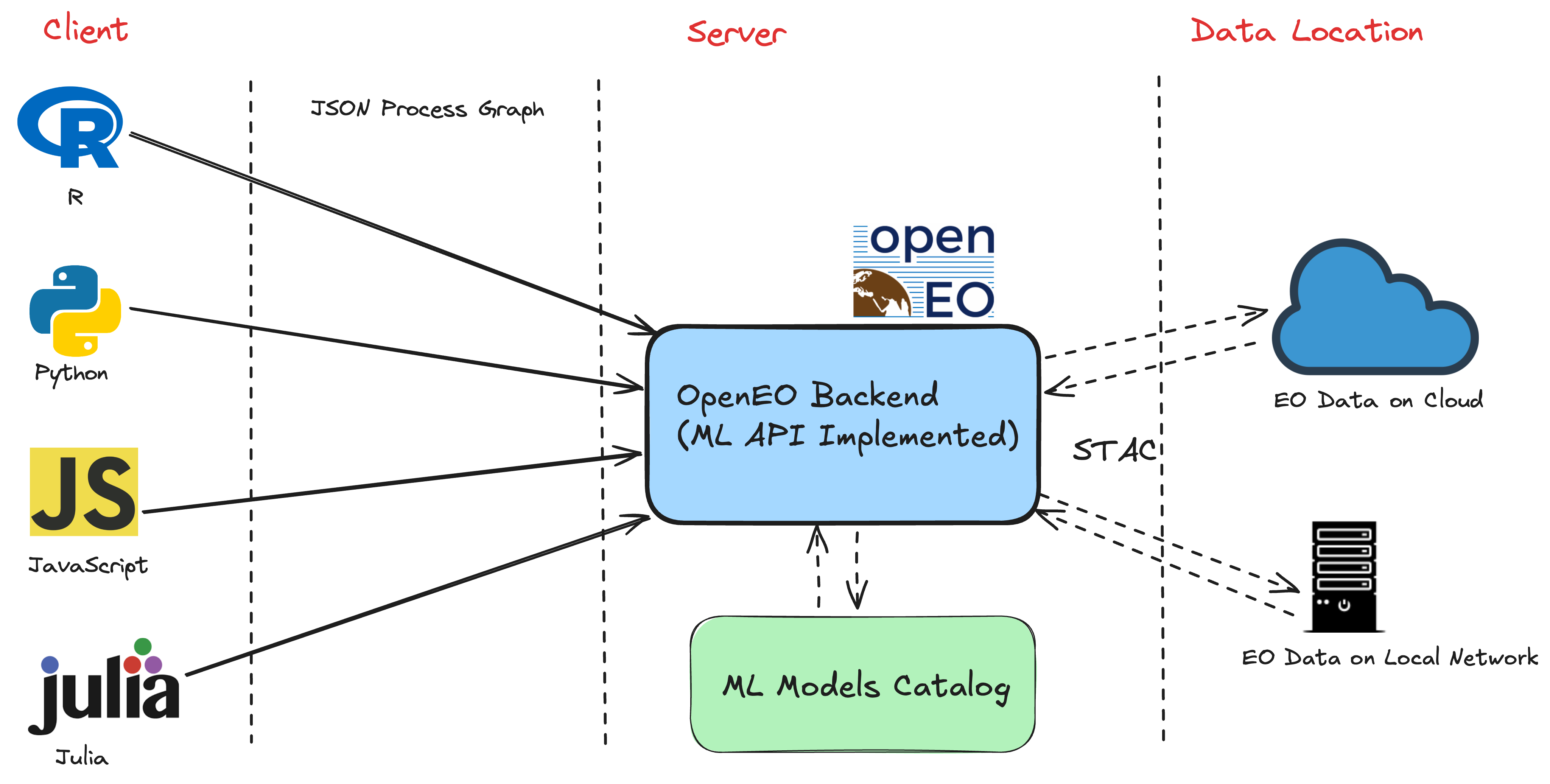}
    \caption{\mbox{openEO} backend interaction with client libraries across
    languages.}
    \label{fig:openEOclients}
\end{figure}

Three prototype implementations of the proposed \mbox{openEO} ML
specification have been developed, each demonstrating how standardized ML
processes can be integrated into EO data cube environments. The
implementations span R and Python ecosystems and cover both classical ML and
deep learning workflows. All three can be used with the \mbox{openEO} client
libraries for R, Python, Julia, and JavaScript, as well as the \mbox{openEO}
web editor (Fig.~\ref{fig:openEOclients}), illustrating the applicability of
the specification across diverse programming environments.

None of the three implementations covers the full ML specification at this
stage; feature completion is a goal for future development.
Table~\ref{tab:implementations} summarizes the capabilities of each
implementation.

\begin{table}[h!]
    \centering
    \caption[Prototype implementation capabilities across openEOcubes, openEOcraft, and openeo-processes-dask-ml]{Overview of the three prototype implementations of the proposed ML
    specification and their respective capabilities. Color coding distinguishes
    the two R-based backends, \mbox{openEOcubes} and \mbox{openEOcraft}
    (blue), from the Python-based \mbox{openeo-processes-dask-ml} backend
    (green). Checkmarks indicate supported features; crosses indicate features
    not yet implemented.}
    \label{tab:implementations}
    \begin{tabular}{l
        >{\columncolor{openeocubes}}c
        >{\columncolor{openeocubes}}c
        >{\columncolor{daskml}}c}
        \toprule[1.2pt]
        & \textbf{openEOcubes}
        & \textbf{openEOcraft}
        & \textbf{openeo-processes-dask-ml} \\
        \midrule[1.2pt]
        Language of implementation  & R              & R              & Python \\
        Full openEO backend         & \cmark         & \cmark         & \xmark \\
        Data cube engine            & gdalcubes      & sits           & xarray \\
        Supported ML frameworks     & R Torch, Caret & R Torch, Caret & PyTorch, Scikit-Learn \\
        Engineered feature vectors  & \cmark         & \cmark         & \cmark \\
        Pixel-wise time series      & \cmark         & \cmark         & \cmark \\
        Spatial neighborhoods       & \xmark         & \xmark         & \cmark \\
        Load pre-trained ML models  & \cmark         & \cmark         & \cmark \\
        Predict with loaded model   & \cmark         & \cmark         & \cmark \\
        Initialize new models       & \cmark         & \cmark         & \cmark \\
        Train new models            & \cmark         & \cmark         & \cmark\ (Random Forest) \\
        Hyperparameter tuning       & \cmark         & \cmark         & \xmark \\
        Save models with metadata   & \cmark         & \cmark         & \cmark \\
        Model saving formats        & ONNX           & R Object       & Python pickle \\
        Model loading formats       & ONNX, .pt      & R Object       & Torchscript, .pt, Python pickle \\
        \bottomrule[1.2pt]
    \end{tabular}
\end{table}
\subsection{openEOcubes}

\mbox{openEOcubes} is an R backend built on top of
\mbox{gdalcubes}~\cite{appel2019}, a library for managing spatiotemporal
raster data as data cubes, exposed through the \mbox{openEO} specification
via a dedicated driver~\cite{pondi2024}. Classical ML workflows are
supported through the \mbox{caret} package~\cite{kuhn2008}. Because ONNX
model export is not natively available in R, the
\mbox{reticulate}\footnote{\url{https://rstudio.github.io/reticulate/}}
package is used to invoke Python tooling for serializing R-trained models to
ONNX format, so that they can be loaded by runtimes outside R. Deep learning
workflows are handled through the
\mbox{torch}\footnote{\url{https://torch.mlverse.org/}} package for R,
which supports training and inference of neural network models within the
\mbox{openEO} process model.

\subsection{openEOcraft}

\mbox{openEOcraft}\footnote{\url{https://github.com/Open-Earth-Monitor/openEOcraft}}
is a generic, open-source R backend designed to expose arbitrary R data cube
implementations as \mbox{openEO}-compliant services~\cite{pondi2026openeocraft}. In contrast to
\mbox{openEOcubes}, which is tightly coupled to \mbox{gdalcubes},
\mbox{openEOcraft} introduces a flexible abstraction layer built around a
lightweight decorator mechanism (\texttt{@openeo-process}) that registers R
functions as \mbox{openEO} processes without requiring changes to the core
API.

This mechanism enables straightforward integration of the
\mbox{\texttt{sits}}~\cite{simoes2021} package, an R library for satellite image
time-series classification, into \mbox{openEO}-compliant workflows.
\mbox{\texttt{sits}} provides a broad range of classical ML and deep learning methods,
and its integration demonstrates how the proposed specification accommodates
diverse modelling approaches across R libraries while remaining consistent
with the \mbox{openEO} process model.

\begin{lstlisting}[language=R, caption={Minimal example of exposing a new
model initialization process via the \texttt{@openeo-process} decorator in
\mbox{openEOcraft}.}, label={lst:decorator-tempcnn}]
#* @openeo-process
mlm_class_tempcnn <- function(
  cnn_layers = c(64, 64, 64),
  cnn_kernels = c(5, 5, 5),
  learning_rate = 5e-4,
  epochs = 150,
  batch_size = 64,
  seed = NULL
) {
  list(
    train = function(training_set) {
      if (!is.null(seed)) set.seed(seed)
      model <- sits::sits_tempcnn(
        cnn_layers = cnn_layers,
        cnn_kernels = cnn_kernels,
        opt_hparams = list(lr = learning_rate),
        epochs = epochs,
        batch_size = batch_size
      )
      sits::sits_train(training_set, model)
    }
  )
}
\end{lstlisting}

Listing~\ref{lst:decorator-tempcnn} illustrates the decorator pattern in
practice. The decorated function is registered as the \mbox{openEO} process
\texttt{mlm\_class\_tempcnn} and becomes immediately accessible to all
\mbox{openEO} client libraries with a well-defined parameter interface. The
function returns a model definition object represented as an R list that
encapsulates the training routine while delegating the concrete
implementation to the \mbox{sits} library. Additional ML methods can be
integrated by adding new decorated functions that follow the same
\texttt{mlm\_} naming conventions and return structure.

openEOcraft's preprocessing is built around \texttt{cube\_regularize}, which wraps \texttt{sits'} \texttt{sits\_regularize} and performs temporal regularization and cloud masking as a single bundled step rather than as separate composable openEO processes. Running the RF workflow in Section~\ref{sec:use-case-1} requires \texttt{aggregate\_temporal\_period} and \texttt{mask} to be exposed individually, which means decomposing sits internals. Since sits is maintained by INPE (Brazil's National Institute
for Space Research) and lies outside the authors' control, this refactoring was not feasible; openEOcraft is therefore excluded from that comparison. It was selected as a prototype backend because \texttt{sits} has been in active development for over five years, tested at continental scale across South America, and covers both classical ML and deep learning in a single package. The sits tibble is the package's native data structure for training
samples and is distributed in most published datasets as RDS files on
GitHub\footnote{\url{https://github.com/e-sensing/sitsdata}}. openEOcraft
accepts these files directly (Section~\ref{sec:use-case-2}), which makes it a
distinctive case for examining how package-specific data formats interact with the specification, a point taken up in Section~\ref{sec:discussion}.

\subsection{\mbox{openeo-processes-dask-ml}}
 
\mbox{openeo-processes-dask-ml}\footnote{\url{https://github.com/Open-EO/openeo-processes-dask-ml}}
extends the \mbox{openeo-processes-dask} library with the ML processes
introduced in this paper. Unlike \mbox{openEOcubes} and
\mbox{openEOcraft}, it is not a standalone \mbox{openEO} backend but a
process implementation library intended for use within existing backends,
such as IBM's
tensorlakehouse-openeo-driver\footnote{\url{https://github.com/IBM/tensorlakehouse-openeo-driver}}
and EODC's
openeo-argoworkflows\footnote{\url{https://github.com/eodcgmbh/openeo-argoworkflows}}.
It uses \mbox{xarray}~\cite{hoyer2017} as its data cube engine and relies on Dask for distributed computation.
 
This is the only implementation among the three that supports all three
EO ML workflow types: engineered feature vectors, pixel-wise time series, and spatial neighborhoods. A key feature is automatic input and output reshaping: the STAC MLM metadata associated with a loaded model specifies the expected input shape and its correspondence to data cube dimensions, and the implementation automatically reshapes the EO data cube to match before inference and reconstructs the model output into a data cube after prediction.
 
At present, the implementation trains Random Forest classification and
regression models through scikit-learn, saves them as Python pickle files
with a STAC MLM Item, and runs inference with pre-trained PyTorch models.
Like other serialized formats that embed executable code, pickle files
should only be loaded from trusted sources or in a sandboxed
environment\footnote{\url{https://docs.python.org/3/library/pickle.html}}.

Adoption by operational backends has been limited by the challenge of robustly integrating parallelized Dask computations with ML frameworks such as PyTorch and CUDA for GPU-accelerated inference. Three issues lie behind this. First, Dask splits
a data cube into chunks and schedules one task per chunk on a pool of
workers~\cite{rocklin2015}. A deep learning model, by contrast, is expensive
to load and should stay in GPU memory for many batches. In our tests,
keeping one model copy per Dask worker, with one GPU per worker, did not
work; the only working configuration inside Dask loaded and unloaded the
model for every chunk, which was very slow. Second, the CUDA runtime cannot
be initialised in a worker process created by copying (forking) its parent
process, so CUDA initialisation has to be placed carefully relative to
worker creation\footnote{\url{https://docs.pytorch.org/docs/stable/notes/multiprocessing.html}}.
Third, the model interface varies. Frameworks built around static dataflow
graphs, such as TensorFlow~\cite{abadi2016}, and exchange formats such as
ONNX~\cite{onnx} describe a model's inputs and outputs as named, typed
tensors, so mapping a chunk of a data cube to the model input is mostly a
matter of reshaping. PyTorch runs models as ordinary Python code, and the
\texttt{forward} method of a module can accept and return arbitrary Python
objects such as dictionaries or lists~\cite{paszke2019}. Automatic
prediction from data cubes therefore needs model-specific glue code. A
Vision Transformer encoder, for instance, returns a list of tensors that each
hold one embedding per image patch, and these have to be reordered into
their spatial layout before they can be written into a cube. We provide this
glue code in a separate package,
\texttt{ml-datacube-bridge}\footnote{\url{https://github.com/jonas-hurst/ml-datacube-bridge}}.

The prototype avoids the first two issues by taking model execution out of
the Dask task graph. Dask computes the preprocessed cube, cuts it into
model-sized batches, and writes each batch to disk as a NumPy array.
Inference then runs in separate Python subprocesses that own the GPUs,
which allows the workload to be distributed over several GPUs with the
model loaded once per GPU. The predictions are written to disk and
assembled back into a data cube by a second Dask step. The cost is extra
disk I/O and the loss of Dask's lazy, chunk-by-chunk evaluation across the
inference step. An operational backend would need a GPU-aware scheduler that
keeps models loaded in GPU memory across jobs. This engineering problem,
rather than the process definitions, is the main obstacle to adoption.
 Section~\ref{sec:use-case-3} demonstrates the current capabilities through a patch-level prediction use case using an EO foundation model.

\subsection{Synthesis Across Implementations}\label{sec:impl-synthesis}

Together, the three implementations show that a standardized ML specification can be realized across multiple \mbox{openEO} backend services and programming environments, for both classical ML and deep learning. They do not yet implement the same subset of processes, which limits how far a single workflow can move between them. Sections~\ref{sec:use-case-1} and~\ref{sec:use-cases-dl} exercise these implementations through feasibility demonstrations. Each use case shows what the specification achieves under realistic conditions and where current limits remain. The federated \mbox{openEO} platform (\url{https://openeo.cloud/}) followed the same path from proof-of-concept~\cite{jacob2021} to matured federation~\cite{mohr2025}.
\section{Cross-Backend Interoperability: Crop Type Mapping in Brittany, France}
\label{sec:use-case-1}

To test whether the proposed \mbox{openEO} ML specification lets one ML workflow run unchanged on independent backends, we designed a crop type mapping use case in which an identical client-side process graph is submitted to R-based \mbox{\texttt{openEOcubes}} and Python-based \mbox{\texttt{openeo-processes-dask-ml}} without modification; openEOcraft is not included in this comparison for reasons detailed in Section~\ref{sec4}. A Random Forest classifier with the same configuration is trained to identify crop types in Brittany (Breizh), France, and applied to map crop types in a neighboring region on each backend independently, allowing the
predictions and evaluation metrics to be directly compared. This use case specifically targets the feature-based workflow type introduced in Section~\ref{sec:workflow_types}, in which a data cube is reduced to a tabular feature matrix before being passed to a classical ML estimator.

\subsection{Workflow Design}
\label{sec:uc1-workflow}

The workflow (Fig.~\ref{fig:openEO-class-rf}) separates a training region
from an independent prediction region. The prediction area is located in
northwestern Brittany, covering approximately $-4.02^{\circ}$ to
$-3.74^{\circ}$ longitude and $48.10^{\circ}$ to $48.20^{\circ}$
latitude. The training area is directly north of it, covering
$-4.02^{\circ}$ to $-3.74^{\circ}$ longitude and $48.20^{\circ}$ to
$48.30^{\circ}$ latitude. Two Sentinel-2 Level 2A data cubes are loaded
for the May to September 2017 growing season, including 12 reflectance
bands and the Scene Classification Layer (SCL), and filtered to scenes
with cloud cover below 50\,\% to reduce noise in the temporal composites.

Both cubes undergo identical preprocessing to ensure consistent feature
construction across training and inference. Cloud and cloud-shadow
masking is derived from the SCL band, after which the SCL band is
dropped and only the reflectance bands are retained. Temporal
aggregation is performed using \texttt{aggregate\_temporal\_period} to
generate cloud-free monthly median composites. The Normalized Difference
Vegetation Index (NDVI) is then derived and appended as an additional
feature band. These steps regularize the original spatio-temporal data
cubes and produce feature-consistent representations suitable for
classical ML.

For training, we use field polygons with crop type labels from the
BreizhCrops dataset~\cite{russwurm2020} for 2017 and clip them to the
bounding box of the training area, yielding 3{,}173 labeled
field-parcel polygons. The \texttt{aggregate\_spatial} operation reduces
the preprocessed training cube to a vector data cube containing, for
each labeled geometry, the median value of every spectral band.

Model initialization follows the standardized pattern introduced in
Section~\ref{sec3}. The classifier is created via
\texttt{mlm\_class\_random\_forest}, specifying 150 trees, square-root
feature sampling at each split, and a fixed random seed for
reproducibility. Training is executed through \texttt{ml\_fit} (invoked
in the Python client via the \texttt{fit()} wrapper), returning a
trained model object. For inference, the trained model is applied to the
independently preprocessed prediction cube using \texttt{ml\_predict}
(via the \texttt{predict()} wrapper), producing a categorical data cube
of crop type predictions over the prediction region. The output is
exported as GeoTIFF using \texttt{save\_result} and submitted as an
asynchronous backend job.

\begin{figure}[H]
\begin{lstlisting}[style=python]
import openeo

# Connect to the openEO backend and authenticate with basic credentials.
connection = openeo.connect(
    url="<host-url>", auth_type="basic",
    auth_options={"username": "<user>", "password": "<password>"}
)

# Define the Sentinel-2 collection, bands, labeled training data.
COLLECTION_NAME = "sentinel-2-l2a"
BAND_NAMES = [
    "coastal", "blue", "green", "red", "rededge1", "rededge2", 
    "rededge3", "nir", "nir08", "nir09", "swir16", "swir22", "scl"
]

# training data
training_data_path = "./breizh_data.geojson"

# Load Sentinel-2 data, mask clouds, build monthly median composites, and add NDVI.
def prepare_datacube(bbox):
    datacube = connection.load_collection(
        collection_id=COLLECTION_NAME,
        spatial_extent=bbox,
        temporal_extent=["2017-05-01T00:00:00Z", "2017-09-30T23:59:59Z"],
        bands=BAND_NAMES,
        max_cloud_cover=50,
    )
    # construct cloud mask
    scl = datacube.band("SCL")
    cloud_mask = (scl == 3) | (scl == 8) | (scl == 9)

    # filter datacube by cloud mask
    datacube = datacube.filter_bands(b for b in BAND_NAMES if b != "SCL")
    datacube = datacube.mask(cloud_mask)

    # build monthly median aggregate
    datacube = datacube.aggregate_temporal_period(period="month", reducer="median")

    # add ndvi
    datacube = datacube.ndvi(red="red", nir="nir", target_band="NDVI")

    return datacube


# Prepare separate datacubes for model training and spatial prediction.
datacube_predict = prepare_datacube({"west": -4.02, "south": 48.10, "east": -3.74, "north": 48.20, "crs": 4326})
datacube_train   = prepare_datacube({"west": -4.02, "south": 48.20, "east": -3.74, "north": 48.30, "crs": 4326})

# aggregate spatial statistics over labeled training geometries
training_data = datacube_train.aggregate_spatial(training_data_path, "median")

# Initialize and train the Random Forest model
rf_model = connection.mlm_class_random_forest(num_trees=150, seed=42, max_variables="sqrt")
rf_model_fitted = rf_model.fit(training_set=training_data, target="class_name")

# Apply the trained model to the prediction datacube
predictions = rf_model_fitted.predict(datacube_predict)

# Save the classified result as GeoTIFF, run the batch job, and download the outputs.
result = predictions.save_result("GTiff")
job = result.create_job()
job.start_and_wait()
job.get_results().download_files("output")
\end{lstlisting}
    \caption{Python \mbox{openEO} client workflow for crop type mapping
    using a Random Forest classifier. The workflow loads Sentinel-2
    data, applies SCL-based cloud masking, selects reflectance bands,
    computes monthly median composites, derives NDVI, aggregates
    training features over labeled geometries, trains the model, and
    applies it to a separate prediction region.}
    \label{fig:openEO-class-rf}
\end{figure}

\subsection{Comparison of Results}
\label{sec:uc1-comparison}

R and Python implementations of Random Forest use different random
number generators and tree-building strategies, so pixel-level agreement
across backends is not expected even with a fixed seed. Following
claim~(ii) (Section~\ref{sec1}), we expect agreement in overall accuracy and
in the dominant class per field. A difference would be concerning if the
backends diverged on classes each handles well independently, or if
spatial disagreement followed geographic rather than classification
boundaries. No formal equivalence test or disagreement threshold was
set.

The most direct test of process graph interoperability is whether two
independent implementations of the same process graph reach comparable
predictive performance. Both backends achieve an overall accuracy of
0.82 on the held-out prediction region (Table~\ref{tab:uc1-results}),
despite relying on different language ecosystems (R with
\texttt{caret} versus Python with \texttt{scikit-learn}) and different
data cube engines (\texttt{gdalcubes} versus \texttt{xarray}). Per-class
metrics follow the same pattern in both implementations: high precision
and recall for corn, wheat, and rapeseed, and substantially lower scores
for permanent and temporary meadows. The remaining numerical differences
are small (typically within $\pm$\,0.05 across all metrics) and are
consistent with implementation-level choices in the underlying ML
libraries that the openEO process graph does not constrain, for
example the specific Random Forest variant, default split criteria,
and internal random-number-generator behavior.

\begin{table}[htbp]
    \centering
    \caption{Per-class precision, recall, and F1 score for the Random
    Forest crop type classifier on the held-out prediction region.}
    \label{tab:uc1-results}
    \begin{tabular}{lcccccc}
        \toprule[1.2pt]
        & \multicolumn{3}{c}{\textbf{openEOcubes}}
        & \multicolumn{3}{c}{\textbf{openeo-processes-dask-ml}} \\
        \cmidrule(lr){2-4} \cmidrule(lr){5-7}
        Overall Accuracy
        & \multicolumn{3}{c}{0.82}
        & \multicolumn{3}{c}{0.82} \\

        \textbf{Class}
            & \textbf{Precision} & \textbf{Recall} & \textbf{F1}
            & \textbf{Precision} & \textbf{Recall} & \textbf{F1} \\
        \midrule[1.2pt]
        Barley            & 0.93 & 0.87 & 0.90 & 0.94 & 0.90 & 0.92 \\
        Corn              & 0.98 & 0.98 & 0.98 & 0.99 & 0.96 & 0.97 \\
        Permanent meadows & 0.65 & 0.63 & 0.64 & 0.70 & 0.56 & 0.62 \\
        Rapeseed          & 0.99 & 0.97 & 0.98 & 1.00 & 1.00 & 1.00 \\
        Temporary meadows & 0.70 & 0.73 & 0.71 & 0.67 & 0.80 & 0.73 \\
        Wheat             & 0.95 & 0.97 & 0.96 & 0.95 & 0.99 & 0.97 \\
        \bottomrule[1.2pt]
    \end{tabular}
\end{table}

Per-class metrics against ground truth do not directly measure agreement
between the two backends. We therefore also compute Cohen's $\kappa$
over 258{,}168 paired pixels at 30\,m resolution, treating the two
backend predictions as independent raters. The backends reach
$\kappa = 0.76$, characterised by Landis and Koch~\cite{landis1977} as
``substantial'' agreement.  We report the point estimate without a confidence interval, since adjacent pixels within fields are not statistically independent~\cite{foody2004}.

\begin{figure}[htbp]
    \centering
    \includegraphics[width=0.7\linewidth]{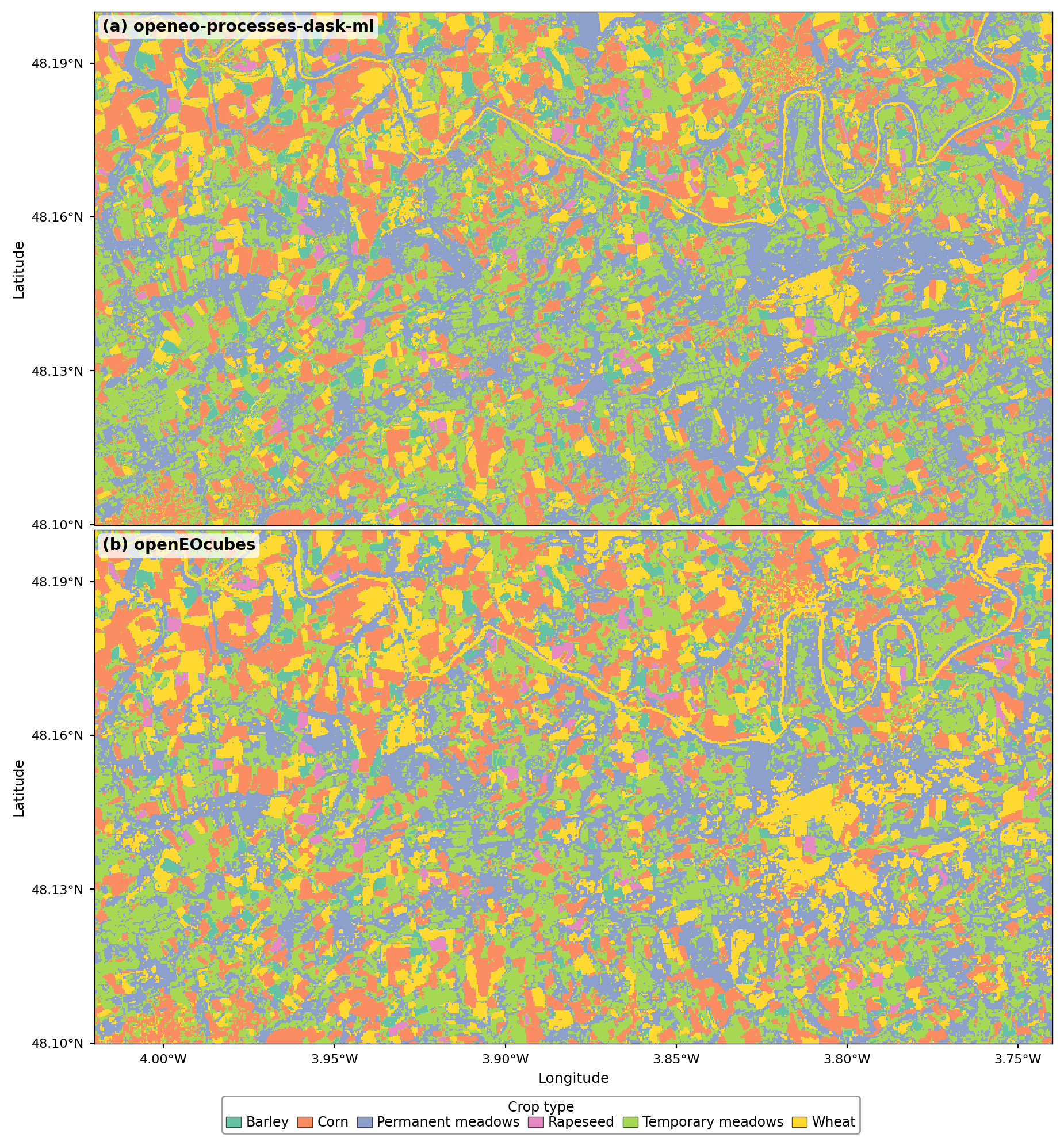}
    \caption{Crop type maps for the prediction region in Brittany
    (Breizh), France, produced by applying the trained Random Forest
    classifier to the Sentinel-2 data cube. The maps show the spatial
    distribution of the six crop classes included in the BreizhCrops
    training dataset, as produced by the
    \texttt{openeo-processes-dask-ml} (top) and \texttt{openEOcubes}
    (bottom) implementations.}
    \label{fig:uc1_class_map}
\end{figure}

\begin{figure}[ht]
    \centering
    \includegraphics[width=0.75\linewidth]{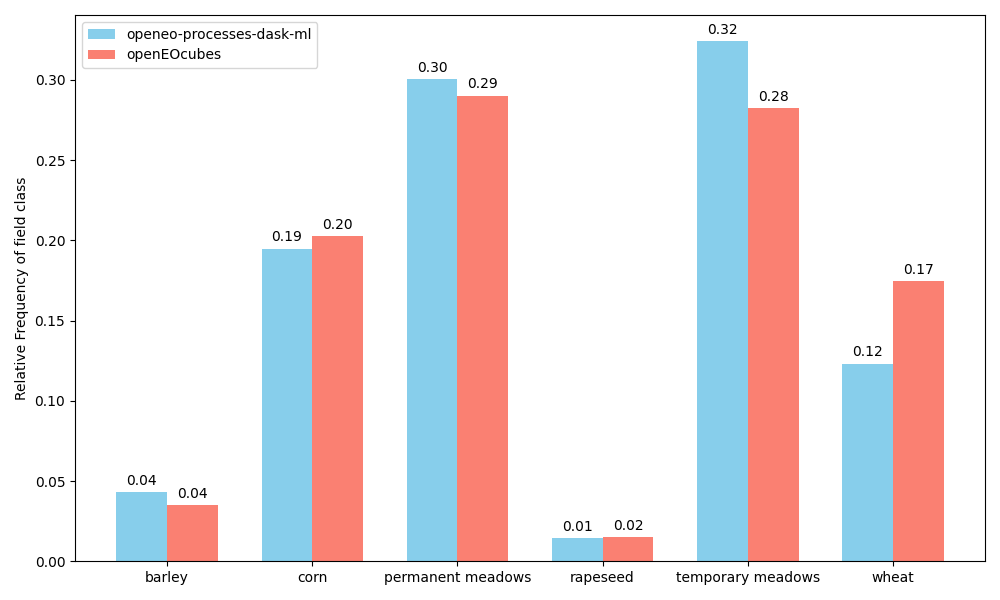}
    \caption{Relative frequency of classes assigned to a pixel in the
    prediction area for the \texttt{openEOcubes} and
    \texttt{openeo-processes-dask-ml} implementations.}
    \label{fig:uc1_histogram}
\end{figure}

Visual inspection of the predicted maps (Fig.~\ref{fig:uc1_class_map})
suggests that the spatial patterns produced by the two backends are
highly consistent at the parcel scale, with minor disagreements
concentrated along field boundaries and in mixed-cover patches. The
class-frequency histogram (Fig.~\ref{fig:uc1_histogram}) shows that the
marginal distributions of predicted classes also align closely between
backends, with the largest discrepancy occurring in the
temporary-meadow / wheat split, which involves the same classes that contribute most
to the lower per-class metrics in Table~\ref{tab:uc1-results}.

Because the prediction area also contains urban surfaces, forests, and
water bodies that the six-class crop classifier was not trained to
recognize, a pixel-wise comparison overstates the relevance of
background disagreements for the agricultural use case. We therefore
complement the pixel-level metrics with a field-level comparison: for
each BreizhCrops field polygon, we extract the modal predicted class
from each backend and compute the proportion of polygons assigned the
same class. Across all 3{,}173 fields, the two backends agree on the
modal crop type for 85\,\% of polygons. The residual disagreements
concentrate on field types where per-class metrics are weakest in
Table~\ref{tab:uc1-results}, notably the permanent-meadow /
temporary-meadow split, consistent with both backends erring on the
same inherently ambiguous cases rather than failing in
implementation-specific ways.

Both expectations stated at the beginning of this section are met, and
neither concerning condition holds. Full bitwise reproducibility still requires deeper
harmonization of ML library defaults and serialization conventions, a
point we return to in Section~\ref{sec:discussion}. Because each
backend trained its own model, the comparison concerns process graph
interoperability. Whether a model trained on one backend can be applied on
the other was not tested: \mbox{openEOcubes} saves Random Forest models as
ONNX files, whereas \mbox{openeo-processes-dask-ml} uses Python pickle files
(Table~\ref{tab:implementations}).

\section{Use Cases: Deep Learning and Foundation Model Workflows}
\label{sec:use-cases-dl}

The following two use cases complement the feature-based Random Forest workflow presented in Section~\ref{sec:use-case-1} by covering the remaining two EO ML workflow types introduced in Section~\ref{sec2}: pixel-wise time series classification and spatial patch-based inference. Together with Section~\ref{sec:use-case-1}, they cover the three workflow types of Section~\ref{sec:workflow_types}, each in a single example.

\subsection{TempCNN for Land Cover Mapping in Brazil}
\label{sec:use-case-2}

This use case shows how the proposed ML API expresses a deep learning workflow for land cover monitoring in Brazil.
The workflow uses a Temporal Convolutional Neural Network
(TempCNN)~\cite{pelletier2019} trained on pixel-level Sentinel-2 time series and executed within an \mbox{openEO}-compliant backend via the \mbox{sits} package~\cite{simoes2021}. It highlights three core aspects of the ML API: model initialization, model tuning/training and prediction, and serialization of model artifacts in the backend.

Training data consist of 6,007 labelled Sentinel-2 time series covering the state of Rondônia in the Brazilian Amazon. Each sample comprises a 16-day composite for 2022 (23 observations) at 10 metres spatial resolution and is assigned to one of nine land cover classes relevant to deforestation monitoring, including \texttt{Clear\_Cut\_Bare\_Soil},
\texttt{Clear\_Cut\_Burned\_Area}, \texttt{Clear\_Cut\_Vegetation}, and
\texttt{Forest}. Samples were collected through visual interpretation and are available in the \texttt{sitsdata} package\footnote{\url{https://github.com/e-sensing/sitsdata}} under a CC-BY 4.0 license.

\begin{figure}[H]
\begin{lstlisting}[language=R,escapeinside={(*@}{@*)}]
# Import relevant libraries/packages
library(openeo)

# Connect to the backend
connection <- connect(host = "<host-url>", user = "<user>", password = "<password>")

# Access processes
p <- processes()

# Training data (public sits samples)
deforestation_data <-
  "https://github.com/e-sensing/sitsdata/raw/main/data/samples_deforestation_rondonia.rds"

# Initialize the TempCNN model (base hyperparameters; grid values override per run)
tempcnn_model_init <- p$mlm_class_tempcnn(
  optimizer = "adam", epochs = 20, batch_size = 64
)

# Hyperparameter search space for fine-tuning
param_grid <- list(
  learning_rate = c(0.0005, 0.0001),
  epochs = c(20, 40)
)

# Grid search: score each combination, then refit on the full training set
tempcnn_tuned <- p$ml_tune_grid(
  model = tempcnn_model_init,
  training_set = deforestation_data,
  target = "label",
  parameters = param_grid,
  scoring = "accuracy",
  cv = 0,
  seed = 42
)

# Save the tuned model; return_model keeps a handle for ml_predict
tempcnn_model <- p$save_ml_model(
  data = tempcnn_tuned,
  name = "tempcnn_rondonia_tuned_v1",
  return_model = TRUE
)

# Load Sentinel-2 data cube (bands must match the training RDS)
datacube <- p$load_collection(
  id = "mpc-sentinel-2-l2a",
  spatial_extent = list(west = -63.50, east = -63.35,
                        south = -8.92, north = -8.78),
  temporal_extent = c("2022-01-01", "2022-12-31"),
  bands = list("B02", "B03", "B04", "B05", "B06", "B07", "B08",
               "B11", "B12", "B8A")
)

# Regularize the data cube to 16-day intervals
datacube <- p$cube_regularize(data = datacube, period = "P16D", resolution = 30)

# Compute NDVI and append as an additional feature band
datacube <- p$ndvi(data = datacube, red = "B04", nir = "B08", target_band = "NDVI")

# Apply the tuned TempCNN to the preprocessed data cube
data <- p$ml_predict(data = datacube, model = tempcnn_model)

# Export the prediction as GeoTIFF and submit as a backend job
ml_job <- p$save_result(data = data, format = "GTiff")
job <- create_job(graph = ml_job,
                  title = "TempCNN fine-tuning + inference")
job <- start_job(job)

# Poll until the job finishes, then retrieve outputs
while (status(job) != "finished") Sys.sleep(30)

# Download the land cover GeoTIFF and tuning report
download_results(job = job, folder = "./results/")
\end{lstlisting}
    \caption{R \mbox{openEO} client code fine-tuning a TempCNN with grid search on labelled Sentinel-2 time series from Rondônia, Brazil, computing NDVI as an additional feature, applying the tuned model for land-cover inference, and downloading the prediction GeoTIFF and tuning report on job completion.}
    \label{fig:openEO-tempcnn}
\end{figure}

Because \mbox{openEOcraft} exposes the standardized ML processes introduced in Section~\ref{sec3}, the entire pipeline is captured in a single process graph (see Appendix~\ref{secA1}) and can be resubmitted unchanged to any backend that implements the same processes and accepts the same training data. At present only \mbox{openEOcraft} does both: the graph uses \texttt{cube\_regularize}, a process built on \texttt{sits}, and references the training samples as a \texttt{sits} RDS file through the string form of \texttt{training\_set} (Section~\ref{sec:model-actions}).

We used the RDS file deliberately: the labelled time series published for
\texttt{sits}, including this set, are distributed in this format, and
accepting it shows that an established EO ML package and its sample
collections can be brought into the openEO ML ecosystem as they are. The
format is readable only from R, which limits interoperability with other
backends (Sections~\ref{sec4} and~\ref{sec:accessibility}). The
\texttt{sits} developers have since added export and import of samples as
Parquet files with GeoParquet metadata, which GDAL, DuckDB, or Python can
read without
\texttt{sits}\footnote{\url{https://github.com/e-sensing/sits/pull/1584}}.
This opens a route to supplying the same samples, as the vector data cube of
Section~\ref{sec:model-actions}, to backends outside R.
 Figure~\ref{fig:openEO-tempcnn} summarises the workflow.

After authenticating with the backend, the client initializes a TempCNN via \texttt{mlm\_class\_tempcnn}, specifying the optimizer and base hyperparameters as provenance metadata. \texttt{ml\_tune\_grid} then conducts a four-run grid search over two learning rates ($0.0005$, $0.0001$) and two epoch counts (20, 40), scoring each candidate by validation accuracy on the Rondônia samples (Table~\ref{tab:tempcnn-grid}). The best configuration (learning rate~$0.0005$, 40~epochs) achieves 93.1\,\% validation accuracy and is refitted on the full training set before being persisted via \texttt{save\_ml\_model}. This artifact can subsequently be retrieved using \texttt{load\_ml\_model} in a new job, so that the same architecture and weights can be reused in later sessions on this backend. Because the artifact is stored as an R object, other backends cannot load it (Table~\ref{tab:implementations}).

\begin{table}[h]
\centering
\caption{TempCNN grid search results on the Rondônia training set.
         Run~2 was selected for final training.}
\label{tab:tempcnn-grid}
\begin{tabular}{cccc}
\hline
Run & Learning rate & Epochs & Validation accuracy (\%) \\
\hline
1 & 0.0005 & 20 & 89.6 \\
2 & 0.0005 & 40 & \textbf{93.1} \\
3 & 0.0001 & 20 & 86.4 \\
4 & 0.0001 & 40 & 88.2 \\
\hline
\end{tabular}
\end{table}

For inference, the workflow loads a Sentinel-2 Level 2A data cube clipped to the Rondônia area of interest for 2022, with spatial $(x, y)$, temporal $(time)$, and band $(band)$ dimensions. Preprocessing is applied within the same process graph: the cube is harmonized to 16-day intervals, matching the training samples, and resampled to a consistent spatial resolution, yielding an aligned multivariate time series per pixel. The \texttt{ndvi} process derives and appends a vegetation index band, producing an augmented feature cube that combines spectral reflectances with a derived index.

\begin{figure}[H]
    \centering
    \includegraphics[width=0.6\textwidth]{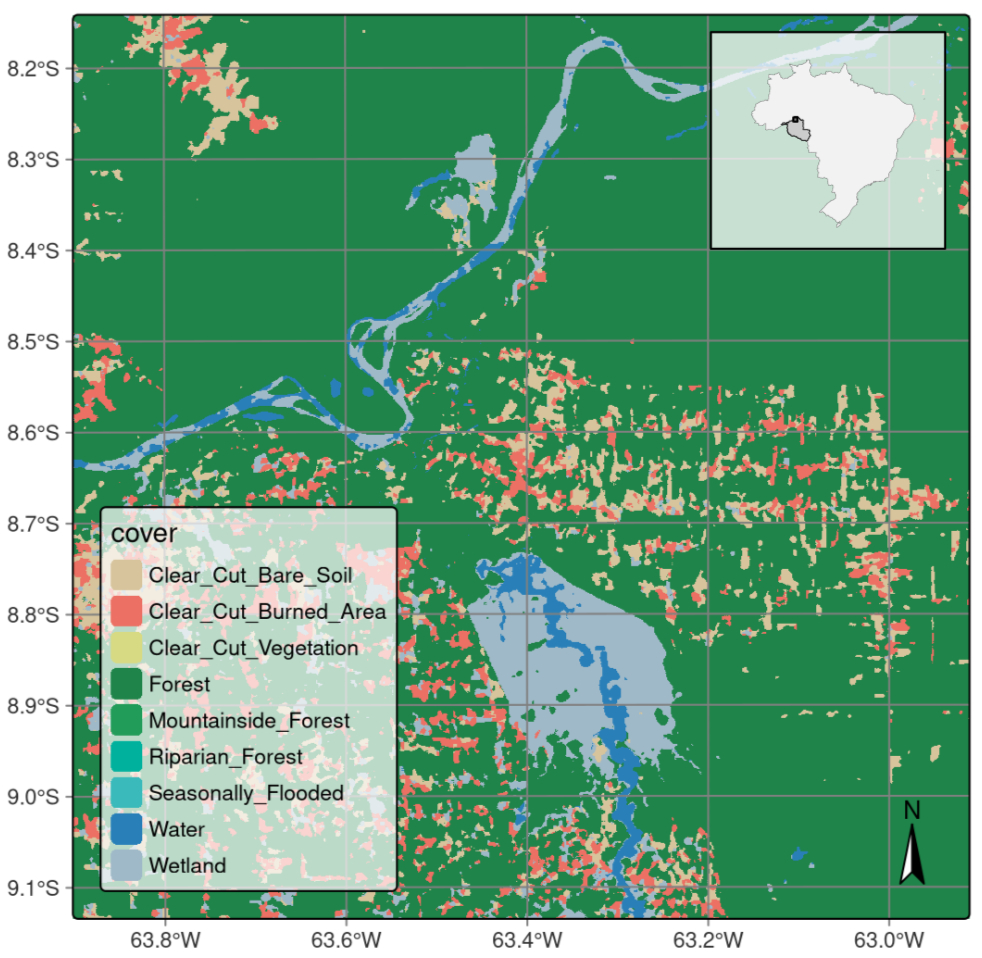}
    \caption{Land cover classification map for the Rondônia study area,
    generated by applying the trained TempCNN model to Sentinel-2 time
    series. The map shows the spatial distribution of the nine land cover
    classes used in training and demonstrates execution of a deep
    learning workflow on EO data cubes, from preprocessing and inference to
    export of a categorical output cube.}
    \label{fig:land-cover}
\end{figure}

The \texttt{ml\_predict} process streams each multiband pixel time series through the trained TempCNN along the temporal dimension, generating a categorical prediction per spatial location. This operation reduces the feature cube to a single classification layer while preserving the spatial grid. The result is exported as GeoTIFF via \texttt{save\_result} and executed asynchronously through \texttt{create\_job} and \texttt{start\_job}.
Once the job completes, \texttt{download\_results} retrieves both the land cover GeoTIFF and the tuning report from the backend. All preprocessing and inference steps remain encoded in the process graph, so the job can be rerun from the graph alone on this backend. The training samples, however, enter as an external file whose own preprocessing is not part of the graph. The exported map is visualized in Fig.~\ref{fig:land-cover}.

\subsection{Foundation Model Inference for EO Embeddings}
\label{sec:use-case-3}

Recent advances in self-supervised representation learning have produced a
growing set of so-called Foundation Models (FM) for EO. These models transform their 
input into compact, task-agnostic embedding vectors which can
generalize to a wide variety of downstream tasks~\cite{jakubik2023}. Many FMs
operate on spatial neighborhoods and are therefore supported only by
the \texttt{openeo-processes-dask-ml} implementation at the time of writing.
As no \mbox{openEO} backend has adopted this implementation yet, client code
cannot be provided. The process graph is instead constructed manually and
invoked as if received through the \mbox{openEO} API.

While the previous use cases (\ref{sec:use-case-1} and \ref{sec:use-case-2}) focused on per-parcel features and pixel-level time series,
we demonstrate in this section how our proposed API supports
spatial patch-based ML tasks. Specifically, we show how \texttt{load\_stac\_ml}
and \texttt{ml\_predict} can be chained to load the pre-trained TerraMind FM~\cite{jakubik2025} from an
external STAC catalog and transform an EO datacube into a datacube of embeddings
through the \mbox{openEO} API.

\begin{figure}[htbp]
    \centering
    \includegraphics[width=1\textwidth]{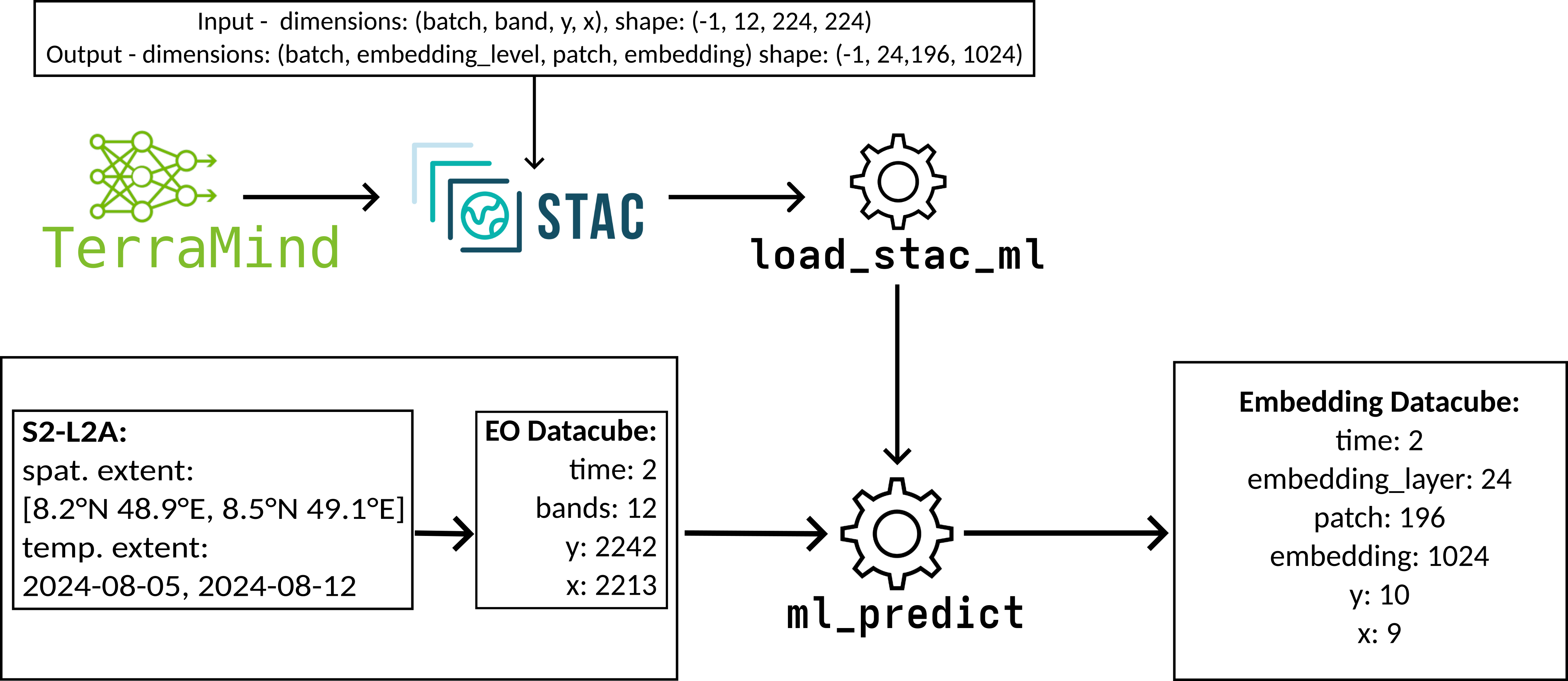}
    \caption{Conceptual chaining of \texttt{load\_stac\_ml} and
    \texttt{ml\_predict} to transform a Sentinel-2 EO data cube into a data
    cube of embeddings using the TerraMind foundation model.}
    \label{fig:ml-predict-fm}
\end{figure}

The pre-trained TerraMind-v1-base backbone is loaded as a
STAC MLM Item~\cite{charette2024} via \texttt{load\_stac\_ml}
(Fig.~\ref{fig:ml-predict-fm}). The STAC MLM metadata specifies the model's
expected input and output shapes and their correspondence to EO data cube
dimensions (the \texttt{batch} dimension is created dynamically and can be of
arbitrary length):

\begin{itemize}
    \item Model input: \texttt{\{batch: -1, band: 12, y: 224, x: 224\}}
    \item Model output: \texttt{\{batch: -1, embedding\_level: 24, patch:
    196, embedding: 1024\}}
\end{itemize}

The \texttt{ml\_predict} process matches model input and output dimension
shapes to the data cube dimensions and transforms them accordingly:

\begin{itemize}
    \item \textbf{time} is present in the data cube but not in the model
    input; the model is therefore applied independently at each time step.

    \item \textbf{band} is present in the data cube and the model input but
    absent from the model output; this dimension is dropped in the output.

    \item \textbf{x, y} are present in the data cube and the model input but
    absent from the model output. The model is applied along the spatial
    dimensions, and each prediction covers the full spatial footprint of the
    input patch. The spatial dimensions in the output data cube are therefore
    reduced such that $y = \lfloor 2242 \div 224 \rfloor = 10$ and $x =
    \lfloor 2213 \div 224 \rfloor = 9$. The remaining $2242 - 10 \times 224
    = 2$ pixels in $y$ and $2213 - 9 \times 224 = 197$ pixels in $x$ are
    insufficient to form a complete input patch and are not represented in
    the output.

    \item \textbf{embedding\_level}, \textbf{patch}, and \textbf{embedding}
    are absent from the model input but present in the model output; these
    dimensions are therefore added to the output data cube.
\end{itemize}

The resulting data cube of embeddings is not visualized here, as embeddings
are an abstract, high-dimensional representation for which direct visual
interpretation is not meaningful. In a next step, an application-specific downstream task model could derive meaningful results from the embeddings (e.g., a water segmentation map), or they could be saved to file and downloaded to the client, which would let the openEO API serve as an interface for on-demand generation of Earth embeddings.

\subsection{Synthesis Across Use Cases}
\label{sec:synthesis-use-cases}

Considered together, the three use cases in Sections~\ref{sec:use-case-1},
\ref{sec:use-case-2}, and~\ref{sec:use-case-3} cover the three 
supervised ML workflow types for EO data cubes identified in
Section~\ref{sec:workflow_types}: feature-based classification using Random
Forests, pixel-wise time-series classification using TempCNN, and spatial
patch-based inference using a geospatial foundation model. In each case, the
same core processes, namely the \texttt{mlm\_} initialization family,
\texttt{ml\_fit} or \texttt{ml\_tune\_*}, \texttt{ml\_predict}, and the model management processes,
are composed within a single declarative process graph. Use Cases~\ref{sec:use-case-1} and~\ref{sec:use-case-2} execute these graphs end-to-end on live
backends; Use Case~\ref{sec:use-case-1} runs across two of the three prototype backends rather than all three, because openEOcraft's bundled
preprocessing step is not directly composable into the process
graph that comparison requires (Section~\ref{sec4}). Use
Case~\ref{sec:use-case-3} demonstrates the dimensional reshaping logic, for inference only and outside a live backend. The use cases therefore show that the three-stage specification can express one example of each of the three workflow types considered here. They do not show that it covers every supervised ML workflow in EO practice, and patch-based training, object-based workflows, and the transfer of trained models between backends remain untested. These limits motivate the reflection on its broader implications that follows.
\section{Discussion}\label{sec:discussion}

Building on the use cases of Sections~\ref{sec:use-case-1} to \ref{sec:use-case-3}, this section reflects on how the proposed specification realizes its stated aims, what the prototype implementations revealed about the practical challenges of
standardizing ML across heterogeneous EO backends, how far it carries over to other datacube standards, and what remains to be addressed in future work.

\subsection{Core Aims and Their Realization}

\subsubsection{Accessibility}\label{sec:accessibility}

By abstracting ML workflows into high-level processes with sensible
default hyperparameters, the specification allows domain experts to
apply methods such as Random Forest, XGBoost, TempCNN, or TAE without
requiring deep programming or ML expertise. Multi-language client
support further lowers the entry barrier by letting users work in their
preferred environment. At the same time, configurability within the
specification is preserved: hyperparameters defined in each process
schema can be explicitly overridden during initialization or
systematically explored through the dedicated tuning processes
\texttt{ml\_tune\_grid} and \texttt{ml\_tune\_random}. Integration with
the STAC MLM extension supports discovery and reuse of pre-trained
models, reducing the computational burden for users who do not need to
train from scratch. The inclusion of validation and tuning processes
extends the specification beyond single-pass model execution, enabling
systematic model development and evaluation within the same declarative
workflow abstraction.

Accessibility depends, however, on more than standardized APIs. The
need to accommodate backend-specific conventions for model serialization
and data handling, including formats such as \mbox{sits} tibbles and R
data frames, shows that further standardization of data exchange
formats across environments remains an open challenge. The range of
models and configurable parameters available in practice also depends on
which processes each backend has implemented; users who require access
to the full parameter space of an underlying library retain more
flexibility by using it directly, and the specification is designed for
interoperability and accessibility rather than as a substitute for direct
library use.

\subsubsection{Interoperability}

The specification can enable executing ML workflows across cloud backends, local installations, and self-hosted systems through backend-agnostic process definitions. The three prototype implementations show that equivalent conceptual workflows can be executed on different infrastructures. Use Case~\ref{sec:use-case-1} confirms this at the process graph level: an identical client-side process graph submitted to both \mbox{openEOcubes} and \mbox{openeo-processes-dask-ml} without modification yields predictions that match in overall accuracy, and converge on the same dominant class at the parcel level with substantial pixel-level agreement ($\kappa = 0.76$), despite the backends relying on entirely different execution engines. Since each backend trained its own model, this is evidence of interoperable process graphs, not of interoperable trained models.

Achieving full interoperability across backends, however, requires alignment
that goes beyond the API layer. Cross-backend execution remains challenging
due to differences in programming language ecosystems, feature engineering
conventions, and model serialization mechanisms. ONNX, as a framework-independent serialization format, can support cross-platform model exchange, but
introduces practical constraints: R backends currently require Python bridging
via the \mbox{reticulate} package to export ONNX artifacts, and models stored
as \mbox{sits}-specific R class objects cannot be loaded by Python
backends. These issues reflect a broader challenge: while the specification
standardizes process interfaces, deeper harmonization of data
formats, serialization standards, and execution semantics is needed before
cross-backend reproducibility can be taken for granted. Unlike a UDF, whose behaviour is defined by user code, an ML process fixes its interface, and the remaining implementation differences are declared through the proposed \texttt{GET /ml\_runtimes} endpoint, already served by \mbox{openEOcubes} and \mbox{openEOcraft} (Section~\ref{sec:capabilities}).

Random Forest was selected as the reference algorithm because it is the
most widely used classifier in  EO workflows~\cite{belgiu2016,maxwell2018} and is available natively across all three prototype backends; the criteria for distinguishing specification failures from implementation-level noise in such a comparison are established in Section~\ref{sec:use-case-1}.

\subsubsection{Reproducibility}

Reproducibility is built into the specification through seed control,
explicit hyperparameter records, and composable process graphs that encode
complete analytical workflows. Use Case~\ref{sec:use-case-2} persists a
tuned model with \texttt{save\_ml\_model} so that it can be reloaded with
\texttt{load\_ml\_model} in later sessions on the same backend. When \texttt{save\_ml\_model} is invoked, the
metadata of Table~\ref{tab:mlm-metadata} is
exported as a STAC MLM Item, providing a structured mechanism for preserving
the computational environment alongside the model artifact.

In practice, reproducibility is still affected by variations in backend
libraries and dependency management. Identical process graphs may yield
slightly different numerical results if model implementations, library versions, preprocessing steps, or data coverage differ across
backends. Handling vector
training labels introduces further ambiguity in cases of partial spatial
overlaps, mixed pixels, or temporal misalignment between features and labels.
Addressing these issues fully will require both tighter conventions for data
exchange and more explicit backend conformance reporting, which the proposed L3-ML profile provides.

\subsubsection{Extensibility}

The modular process structure provides a foundation for community-driven
extensions that can accommodate emerging ML paradigms without altering the core specification. The integration of the TerraMind foundation
model~\cite{jakubik2025} in Use Case~\ref{sec:use-case-3} demonstrates how large pre-trained models can be incorporated into the same standardized interface for inference; downstream adaptation through the same interface has not been demonstrated yet. Foundation models such as TESSERA~\cite{feng2026}, Prithvi~\cite{szwarcman2025} and TerraMind~\cite{jakubik2025}, trained on large EO datasets, offer the possibility of reducing labeled data requirements and improving model generalization across regions and tasks. 

Federated learning workflows~\cite{morenoalvarez2024} represent another candidate for incremental integration through the same process-level abstraction. Random Forest is a natural starting point: local forests trained on regional data cubes at each backend can be retrieved via
\texttt{load\_stac\_ml} and aggregated into a global ensemble~\cite{hauschild2022,liu2020}, with no raw training data
leaving any backend.

Long-term success will require sustained community participation, consistent governance, and iterative refinement based on operational feedback. The \mbox{openEO} ecosystem supports these goals through its Project Steering Committee and regular community meetings, which provide structured mechanisms for shared decision-making and transparent evolution of the specification.

\subsection{Lessons from Prototype Implementations}

The three prototype implementations reveal how backend-specific design
choices interact with the shared specification and where the boundaries of interoperability currently lie.

\mbox{openEOcubes}, built on \mbox{gdalcubes}, R Torch, and \mbox{caret}, supports both feature-based and time-series workflows but requires custom ONNX export pathways that depend on Python bridging, introducing a fragility that would not arise in a natively cross-language serialization standard.
\mbox{openEOcraft} supports both workflow types, but as discussed in
Section~\ref{sec4}, its bundled preprocessing step, and its \texttt{sits}-native data model constrain interoperability with other backends. The Python implementation using \mbox{xarray} and Dask enables distributed computation and is the only implementation covering all three workflow types, but it introduces different constraints around memory management, process scheduling, and GPU integration that have so far limited adoption by operational backends (Section~\ref{sec4}).

Together, these differences illustrate both the value and the limits of a shared specification. The specification defines a consistent conceptual model that allows users to express ML workflows in a backend-agnostic form. What it cannot yet fully guarantee is that those workflows execute identically across backends, because execution fidelity depends on decisions made below the process level: library versions, serialization formats, and data structure conventions that fall outside the scope of the current specification. The proposed L3-ML profile in Appendix~\ref{secD1} is a first step toward making these decisions explicit, since it declares frameworks, versions, and model formats per backend. Recording known behavioural differences, such as default hyperparameters, and adding automated conformance tests are the next steps.

\subsection{Applicability Beyond openEO}\label{sec:beyond-openeo}

openEO is not the only interface to EO data cubes. The Open Geospatial Consortium (OGC) and International Organization for Standardization (ISO) coverage
standards define a data model for multidimensional grids, the Coverage
Implementation Schema (CIS), services to access coverages (WCS), and a
query language for server-side processing, the Web Coverage Processing
Service (WCPS)~\cite{baumann2010,baumann2025}. WCPS expresses processing in
a query language rather than as a catalog of separately specified
processes, so an ML model would have to enter it as a new built-in function
or an external call, neither of which is standardized. Porting our
specification to WCS and WCPS would thus require extending those standards,
although the dimension-matching rules of \texttt{ml\_predict}
(Section~\ref{sec:use-case-3}) carry over to any coverage with named axes.

The fit is closer for OGC API Processes~\cite{pross2021}, which publishes
processes together with machine-readable descriptions of their inputs and
outputs, and for the draft OGC API GeoDataCube (GDC) standard developed in
OGC Testbed-19, which adopts openEO process graphs and names their
submission to OGC API Processes endpoints as future
work~\cite{jacob2024ogc}. A GDC implementation that accepts openEO process
graphs could offer our processes without redefinition; exposing them
individually through OGC API Processes would require translating the
\texttt{datacube} and \texttt{ml-model} types, which we have not attempted.

\section{Conclusion}\label{sec6}

This paper introduced a standardized ML specification for EO data cubes
within the \mbox{openEO} ecosystem, organized into three coherent stages:
model initialization, model actions, and model management. Three prototype
implementations in R and Python, exercised through crop type mapping,
land cover classification, and foundation model inference use cases,
demonstrate that the specification is practically realizable across diverse
technology stacks and that an identical process graph submitted to
independent backends yields predictions that match in overall accuracy and
converge on the same dominant class at the parcel level, with each backend training its own model.

The use cases collectively cover all three EO ML workflow types identified
in this paper: feature-based classification, pixel-wise time-series
modeling, and spatial patch-based inference. This shows that
a single process-level abstraction can express one example of each of these
workflow types; whether it suffices for other paradigms has not been
tested. The integration of the TerraMind foundation
model~\cite{jakubik2025} and compliance with the STAC MLM
extension~\cite{charette2024} further show that the specification
can accommodate pre-trained model reuse and catalog-based model
discovery, reducing the barrier to entry for users who do not need to
train from scratch.

The specification gives ML-based processing of EO data cubes a common, declarative form, but full interoperability across backends requires deeper
alignment of data formats, model serialization standards, and execution
semantics than the current process-level specification alone can enforce.
Reproducibility is still affected by backend library versions and
preprocessing conventions that fall outside the specification boundary.
Addressing these gaps systematically, through explicit backend conformance
profiles and tighter data exchange conventions, is the most important
direction for near-term development. The proposed L3-ML profile (Appendix~\ref{secD1}) is a first step. A train-once, load-elsewhere experiment, in which a model trained on one backend is applied on another, is the next step toward interoperable trained models; ONNX is a candidate format for it, provided exports are verified numerically~\cite{pondi2026onnx}.

The modular architecture provides a foundation for incremental extension.
The decorator-based registration mechanism demonstrated in
\mbox{openEOcraft} shows how new algorithms can be integrated without
modifying the core API, and the same process-level abstraction provides
a foundation for future extension to emerging paradigms, including
federated learning~\cite{morenoalvarez2024} and self-supervised
foundation models~\cite{jakubik2023,szwarcman2025, feng2026}. Operational adoption
in environments such as the Copernicus Data Space Ecosystem, the federated
\mbox{openEO} platform, VITO's Terrascope platform, and IBM's
TensorLakeHouse is a natural next step, contingent on solving the GPU
scheduling problems between distributed computation frameworks and ML
runtimes described in Section~\ref{sec4}.

By defining process-level abstractions that bridge EO data cubes with
tabular and tensor-based ML representations, the specification is a step
toward ML workflows on EO data that are interoperable, reproducible, and
accessible across the infrastructure heterogeneity that characterizes operational
Earth observation today.
\section{Availability and Requirements}\label{sec7}

\textbf{ML API Specification:}
The proposed ML API specification is openly available under the Apache
License Version 2.0 in the \texttt{proposals} directory of the following
repository:
\url{https://github.com/PondiB/openeo-processes/tree/ml-api}.

\textbf{Use Case Examples:}
Reproducible code for all use cases presented in this paper, including
the crop type mapping workflow (Section~\ref{sec:use-case-1}), the TempCNN land
cover classification (Section~\ref{sec:use-case-2}), and the foundation
model embedding pipeline (Section~\ref{sec:use-case-3}), is openly
available under the Apache License Version 2.0 at:
\url{https://github.com/PondiB/openeo-ml-showcase}.

\textbf{openEOcubes:}
Source code is available under the Apache License Version 2.0 at:
\url{https://github.com/PondiB/openeocubes}.

\textbf{openEOcraft:}
Source code is available under the MIT License at:
\url{https://github.com/Open-Earth-Monitor/openeocraft}.

\textbf{openeo-processes-dask-ml:}
Source code is available under the Apache License Version 2.0 at:
\url{https://github.com/Open-EO/openeo-processes-dask-ml}.

\textbf{ml-datacube-bridge:}
Glue code for converting between data cube subsets and the inputs and
outputs of PyTorch models, used by \mbox{openeo-processes-dask-ml}, is
available at: \url{https://github.com/jonas-hurst/ml-datacube-bridge}.

\textbf{Operating System:}
All three prototype implementations are compatible with Windows, macOS,
and Linux.

\section*{Acknowledgments}
We gratefully acknowledge financial support from the European Union's
Horizon Europe research and innovation programme. Authors BP and JS are
supported under agreement No.~101059548 for the Open-Earth-Monitor
Cyberinfrastructure (OEMC) project and under agreement No.~101058386 for
the interdisciplinary Digital Twin Engine for Science (interTwin) project.
Author JH is supported under agreement No.~101131841 for the Embed2Scale
project. Additional funding was provided by the Swiss State Secretariat
for Education, Research and Innovation and by UK Research and Innovation.

\vspace{0.5cm}
\textbf{Authors' contributions:}
\textbf{Brian Pondi:} Writing -- Original Draft, Conceptualization,
Software, Methodology, Visualization, Validation, Investigation.
\textbf{Jonas Hurst:} Writing -- Original Draft, Formal Analysis, Software, Validation,
Visualization, Investigation.
\textbf{Rolf Simoes:} Writing -- Original Draft, Software, Validation,
Visualization, Investigation.
\textbf{Jonas Starke:} Software, Visualization, Data Curation,
Investigation.
\textbf{Marius Appel:} Writing -- Review and Editing, Supervision.
\textbf{Edzer Pebesma:} Writing -- Review and Editing, Supervision,
Resources, Project Administration.

\vspace{0.5cm}
\textbf{Data Availability:}
The BreizhCrops benchmark dataset is openly available from
\cite{russwurm2020} at \url{https://breizhcrops.org/}. The Rondônia
deforestation dataset is provided by Brazil's National Institute for
Space Research (Instituto Nacional de Pesquisas Espaciais) at
\url{https://github.com/e-sensing/sitsdata}.

\vspace{0.5cm}
\textbf{Funding:}
This work was supported by the European Commission under Horizon Europe
agreement No.~101059548 (Open-Earth-Monitor Cyberinfrastructure),
No.~101058386 (interdisciplinary Digital Twin Engine for Science), and
No.~101131841 (Embed2Scale), as well as by the Swiss State Secretariat
for Education, Research and Innovation, and by UK Research and Innovation.

\vspace{0.5cm}
\textbf{Competing Interests:}
The authors declare no competing interests.

\newpage
\begin{appendices}
  \section{ML Process Graph}\label{secA1}

This appendix provides a concrete example of an \mbox{openEO} JSON process
graph generated from the ML workflow described in
Section~\ref{sec:use-case-2}. The graph captures all data cube
preprocessing, TempCNN hyperparameter fine-tuning, model export, and inference
steps required to execute the workflow on a backend
(Fig.~\ref{fig:openEO-json-graph}). By expressing these steps
declaratively, the process graph makes the sequence of operations, data
dependencies, and model interactions explicit and reproducible.

\begin{figure}[H]
\begin{lstlisting}
{
 "process_graph": {
  "loadcollection1": {
   "process_id": "load_collection",
   "arguments": {
    "id": "mpc-sentinel-2-l2a",
    "spatial_extent": {"west":-63.50,"south":-8.92,"east":-63.35,"north":-8.78},
    "temporal_extent": ["2022-01-01","2022-12-31"],
    "bands": ["B02","B03","B04","B05","B06","B07","B08","B11","B12","B8A"]
   }
  },
  "cuberegularize1": {
   "process_id": "cube_regularize",
   "arguments": {"data":{"from_node":"loadcollection1"},"period":"P16D","resolution":30}
  },
  "ndvi1": {
   "process_id": "ndvi",
   "arguments": {"data":{"from_node":"cuberegularize1"},"red":"B04","nir":"B08","target_band":"NDVI"}
  },
  "mlmclasstempcnn1": {
   "process_id": "mlm_class_tempcnn",
   "arguments": {"optimizer":"adam","learning_rate":0.0005,"epochs":20,"batch_size":64}
  },
  "mltunegrid1": {
   "process_id": "ml_tune_grid",
   "arguments": {
    "model":{"from_node":"mlmclasstempcnn1"},
    "training_set":"{<deforestation_data>}",
    "target":"label",
    "parameters":{"learning_rate":[0.0005,0.0001],"epochs":[20,40]},
    "scoring":"accuracy","cv":0,"seed":42
   }
  },
  "savemlmodel1": {
   "process_id": "save_ml_model",
   "arguments": {"data":{"from_node":"mltunegrid1"},"name":"tempcnn_rondonia_tuned_v1","return_model":true}
  },
  "mlpredict1": {
   "process_id": "ml_predict",
   "arguments": {"data":{"from_node":"ndvi1"},"model":{"from_node":"savemlmodel1"}}
  },
  "saveresult1": {
   "process_id": "save_result",
   "arguments": {"data":{"from_node":"mlpredict1"},"format":"GTiff"},
   "result": true
  }
 }
}
\end{lstlisting}
    \caption{openEO JSON process graph for the TempCNN fine-tuning and land cover
    mapping workflow. Each node in \texttt{process\_graph} corresponds to one
    \mbox{openEO} process, with \texttt{from\_node} references expressing
    the data flow between operations.}
    \label{fig:openEO-json-graph}
\end{figure}

Each node in the graph corresponds to a single \mbox{openEO} process, and
data flow between nodes is expressed through \texttt{from\_node} references
rather than intermediate variables, making dependencies explicit.
The \texttt{load\_collection}, \texttt{cube\_regularize}, and \texttt{ndvi}
nodes define the preprocessing applied to the Sentinel-2 data cube before
inference. The \texttt{mlm\_class\_tempcnn} node declares the base TempCNN
architecture and hyperparameters; \texttt{ml\_tune\_grid} evaluates a
parameter grid on the labelled training samples, selects the best
configuration, and refits the model on the full training set (writing
\texttt{tuning\_results.json} in the job workspace). The
\texttt{save\_ml\_model} node persists the tuned model while
\texttt{return\_model: true} keeps it available as an input to downstream
processes. \texttt{ml\_predict} applies the tuned model to the preprocessed
cube, and \texttt{save\_result} exports the classification as a GeoTIFF.
The \texttt{result: true} flag on the final node signals to the backend which
output to return when the job completes.
  \newpage
  \section{Example ML Process Schema}\label{secB1}

This appendix shows an example JSON process specification for
\texttt{ml\_predict} (Fig.~\ref{fig:ml-predict-specification}), illustrating how
ML processes are formally defined within the \mbox{openEO} process catalog.
Process specifications define the expected inputs, outputs, and data types,
providing a shared contract that all backend implementations must satisfy.

\begin{figure}[H]
\begin{lstlisting}
{
  "id": "ml_predict",
  "summary": "Apply a trained ML model to a data cube",
  "description": "Applies a machine learning model to a data cube of
    input features and returns the predicted values.",
  "categories": ["machine learning"],
  "experimental": true,
  "parameters": [
    {
      "name": "data",
      "description": "The data cube containing the input features.",
      "schema": {
        "type": "object",
        "subtype": "datacube"
      }
    },
    {
      "name": "model",
      "description": "A trained ML model, as returned by ml_fit or
        loaded via load_ml_model or load_stac_ml.",
      "schema": {
        "type": "object",
        "subtype": "ml-model"
      }
    }
  ],
  "returns": {
    "description": "A data cube with the predicted values. The input
      feature dimensions are removed and a new dimension named
      'predictions' of type 'other' is added. If a single value is
      returned per location, this dimension carries a single label '0'.",
    "schema": {
      "type": "object",
      "subtype": "datacube",
      "dimensions": [{"type": "other"}]
    }
  }
}
\end{lstlisting}
    \caption{JSON process specification for \texttt{ml\_predict}, defining
    the expected input types, output structure, and data cube transformation
    behavior. The \texttt{experimental} flag indicates that the process
    definition is subject to revision as the specification matures.}
    \label{fig:ml-predict-specification}
\end{figure}

The specification follows the same structural pattern as all other
\mbox{openEO} process definitions, ensuring that \texttt{ml\_predict} is
treated as a first-class citizen in the process catalog rather than a
special-case extension. Both parameters are typed as \mbox{openEO} objects:
\texttt{data} must be a data cube, and \texttt{model} must be an
\texttt{ml-model} object produced by \texttt{ml\_fit},
\texttt{load\_ml\_model}, or \texttt{load\_stac\_ml}. The return type is
also a data cube, which means the output of \texttt{ml\_predict} can be
passed directly to any subsequent \mbox{openEO} process, such as
\texttt{ml\_uncertainty\_class}, \texttt{ml\_label\_class}, or
\texttt{save\_result}, without requiring format conversion. The
\texttt{experimental} flag signals that the definition may be refined
through community feedback before it is promoted to a stable process.
  \newpage
  \section{openEO Workflow for Foundation Model Embeddings}\label{secC1}

This appendix provides the \mbox{openEO} process graph corresponding to the
embedding pipeline described in Section~\ref{sec:use-case-3}
(Fig.~\ref{fig:openeo-embeddings-client}). The workflow loads Sentinel-2 data as a
data cube, retrieves the pre-trained TerraMind foundation model described as
a STAC MLM Item, and applies the model to compute embeddings. The output is
a high-dimensional data cube intended for downstream processing rather than
immediate export.

\begin{figure}[H]
\begin{lstlisting}
{
 "process_graph": {
  "load_data": {
   "process_id": "load_stac",
   "arguments": {
    "url": "https://earth-search.aws.element84.com/v1/collections/sentinel-2-l2a",
    "spatial_extent": {"west": 8.2, "east": 8.5, "south": 48.9, "north": 49.1},
    "temporal_extent": ["2024-08-05", "2024-08-12"],
    "bands": [
     "coastal", "blue", "green", "red",
     "rededge1", "rededge2", "rededge3",
     "nir", "nir08", "nir09", "swir16", "swir22"
    ],
    "resolution": 10
   }
  },
  "load_model": {
   "process_id": "load_stac_ml",
   "arguments": {
    "uri": "<url-to-stac-mlm-item>",
    "model_asset": "weights"
   }
  },
  "predict": {
   "process_id": "ml_predict",
   "arguments": {
    "data": {"from_node": "load_data"},
    "model": {"from_node": "load_model"}
   },
   "result": true
  }
 }
}
\end{lstlisting}
    \caption{openEO process graph for computing EO embeddings from
    Sentinel-2 data using the TerraMind foundation model. The workflow
    loads an EO data cube, retrieves a pre-trained model via
    \texttt{load\_stac\_ml}, and applies inference via \texttt{ml\_predict}.
    The resulting embedding cube is an intermediate output intended for
    further downstream processing.}
    \label{fig:openeo-embeddings-client}
\end{figure}

In contrast to the supervised ML pipelines in Sections~\ref{sec:use-case-1} and~\ref{sec:use-case-2}, this
workflow contains no \texttt{ml\_fit} step, as the foundation model is
pre-trained and used exclusively for inference. The three steps are
therefore: (i)~EO data loading via \texttt{load\_stac}, (ii)~model
retrieval via \texttt{load\_stac\_ml}, and (iii)~embedding generation via
\texttt{ml\_predict}. The resulting embedding cube can subsequently be
passed to further \mbox{openEO} processes within the same declarative
workflow, for example for dimensionality reduction, clustering, or
fine-tuning on a downstream classification task.

  \section{Proposed L3-ML Capability Profile}\label{secD1}

openEO describes what a backend supports through process profiles and API
profiles\footnote{\url{https://openeo.org/documentation/1.0/developers/profiles/index.html}}.
This appendix proposes both parts for ML. Neither is ratified; they will be
proposed to the openEO Project Steering Committee together with the process
definitions of Section~\ref{sec3}. The goal is that a client, or a broker in
a federation~\cite{mohr2025}, can answer three questions before it submits a
job: can this backend train the requested model type, can it load the model
format at hand, and can it run the required workflow type.

\subsection{Process profile}

The existing L3-ML process profile of openEO lists a few legacy ML
processes. We propose to extend it with the processes of Section~\ref{sec3} and
to split it into two levels, so that backends that only run pre-trained
models can still declare a meaningful profile.

\begin{table}[h!]
    \centering
    \caption{Proposed levels of the L3-ML process profile.}
    \label{tab:l3ml-process}
    \begin{tabular}{@{\hspace{0.5em}}p{0.22\textwidth}@{\hspace{0.02\textwidth}}p{0.70\textwidth}@{\hspace{0.5em}}}
        \toprule[1.2pt]
        \textbf{Level} & \textbf{Required processes} \\
        \midrule[1pt]
        L3-ML Inference & \texttt{load\_stac\_ml}, \texttt{ml\_predict} \\
        \addlinespace[0.3em]
        L3-ML Training & All of L3-ML Inference, plus at least one
        \texttt{mlm\_<type>\_<model>} process, \texttt{ml\_fit}, 
        \texttt{save\_ml\_model} and \texttt{load\_ml\_model} \\
        \addlinespace[0.3em]
        Optional & \texttt{ml\_predict\_probabilities},
        \texttt{ml\_tune\_grid}, \texttt{ml\_tune\_random},
        \texttt{ml\_validate}, \texttt{ml\_validate\_kfold},
        \texttt{ml\_uncertainty\_class}, \texttt{ml\_smooth\_class},
        \texttt{ml\_label\_class} \\
        \bottomrule[1.2pt]
    \end{tabular}
\end{table}

\subsection{API profile}

The API part follows the L3-UDF profile, where \texttt{GET /udf\_runtimes}
lists the available UDF runtimes and their versions. We propose two
endpoints with the requirements in Table~\ref{tab:l3ml-api}:
\texttt{GET /ml\_runtimes} declares the ML runtimes of a backend, and
\texttt{GET /ml\_models} lists the models stored on it, analogous to
\texttt{GET /collections} for \texttt{load\_collection}. Requirement identifiers use the prefix ML to
avoid clashes with the numbering of the existing openEO profiles. Both
endpoints are proposals at the level of the openEO API; \mbox{openEOcubes}
and \mbox{openEOcraft} already serve \texttt{GET /ml\_runtimes}.

\begin{table}[h!]
    \centering
    \caption{Proposed requirements of the L3-ML API profile.}
    \label{tab:l3ml-api}
    \begin{tabular}{@{\hspace{0.5em}}p{0.07\textwidth}@{\hspace{0.02\textwidth}}p{0.25\textwidth}@{\hspace{0.02\textwidth}}p{0.58\textwidth}@{\hspace{0.5em}}}
        \toprule[1.2pt]
        \textbf{\#} & \textbf{Functionality} & \textbf{Description} \\
        \midrule[1pt]
        ML1 & \texttt{GET /ml\_runtimes} & Is supported and listed in the \texttt{endpoints} of \texttt{GET /} \\
        ML2 & \texttt{GET /ml\_runtimes} & Works with and without authentication \\
        ML3 & \texttt{GET /ml\_runtimes} & Valid response with at least \texttt{default} and \texttt{versions} per framework; framework names follow the STAC MLM \texttt{mlm:framework} vocabulary \\
        ML4 & \texttt{GET /ml\_runtimes} $>$ versions & States per version whether training and inference are supported \\
        ML5 & \texttt{GET /ml\_runtimes} $>$ versions & Lists the \texttt{mlm:artifact\_type} values that can be loaded and saved \\
        ML6 & \texttt{GET /ml\_runtimes} $>$ versions & Lists the supported workflow types (\texttt{feature}, \texttt{time\_series}, \texttt{spatial\_patch}) \\
        ML7 & \texttt{GET /ml\_runtimes} $>$ versions & Lists the accepted training data formats; the vector data cube is always listed when training is supported \\
        ML8 & \texttt{GET /ml\_runtimes} $>$ versions & Lists the available accelerators using the STAC MLM \texttt{mlm:accelerator} vocabulary \\
        ML9 & Errors & Mismatches between model metadata and runtime are reported as described in Section~\ref{sec:model-management} and name the offending field \\
        ML10 & \texttt{GET /ml\_models} & Is supported, listed in the \texttt{endpoints} of \texttt{GET /}, and lists the models stored on the backend with their STAC MLM metadata \\
        ML11 & \texttt{GET /ml\_models/\{id\}} & Returns the full STAC MLM Item of one model; the identifier can be passed to \texttt{load\_ml\_model} \\
        \bottomrule[1.2pt]
    \end{tabular}
\end{table}

\subsection{Profile of the prototype implementations}

Table~\ref{tab:l3ml-prototypes} applies the proposed profile to the three
prototypes as described in Section~\ref{sec4}. It also makes visible which
combinations are not yet possible: no two prototypes share a model format
that one writes and the other reads, and only
\mbox{openeo-processes-dask-ml} supports spatial patches.

\begin{table}[h!]
    \centering
    \caption{Proposed L3-ML profile applied to the three prototype implementations.}
    \label{tab:l3ml-prototypes}
    \begin{tabular}{@{\hspace{0.3em}}p{0.20\textwidth}p{0.22\textwidth}p{0.22\textwidth}p{0.26\textwidth}@{\hspace{0.3em}}}
        \toprule[1.2pt]
        & \textbf{openEOcubes} & \textbf{openEOcraft} & \textbf{openeo-processes-dask-ml} \\
        \midrule[1pt]
        \texttt{GET /ml\_runtimes} & Served & Served & Depends on the host backend \\
        \texttt{GET /ml\_models} & Served & Served & Depends on the host backend \\
        Process level & Training & Training & Training (Random Forest only) \\
        Frameworks & caret, torch (R) & caret, torch (R) & scikit-learn, PyTorch \\
        Load formats & ONNX, \texttt{.pt} & R object & TorchScript, \texttt{.pt}, Python pickle \\
        Save formats & ONNX & R object & Python pickle \\
        Workflow types & feature, time series & feature, time series & feature, time series, spatial patch \\
        Extra training data formats & none & \texttt{sits} tibble (RDS) & none \\
        \bottomrule[1.2pt]
    \end{tabular}
\end{table}

\end{appendices}

\newpage
\bibliographystyle{sn-mathphys-num}
\bibliography{sn-bibliography}

\end{document}